\documentclass[10pt,twocolumn,letterpaper]{article}

\usepackage{iccv} %
\usepackage{graphicx}
\usepackage{amsmath}
\usepackage{amssymb}
\usepackage{booktabs}
\usepackage{multirow}
\usepackage{xcolor}

\usepackage[accsupp]{axessibility}  %

\definecolor{iccvblue}{rgb}{0.21,0.49,0.74}
\usepackage[pagebackref,breaklinks,colorlinks,allcolors=iccvblue]{hyperref}

\definecolor{pastelblue}{RGB}{173,216,230}
\definecolor{pastelred}{RGB}{255,182,193}
\definecolor{pastelgreen}{RGB}{144,238,144}
\definecolor{pastelorange}{RGB}{255,222,173}
\definecolor{pastelpurple}{RGB}{216,191,216}

\definecolor{violett}{RGB}{151, 115, 166}
\definecolor{grenn}{RGB}{129, 179, 101}

\newcommand{\dataname}{SiM3D}

\usepackage[capitalize]{cleveref}
\crefname{section}{Sec.}{Secs.}
\Crefname{section}{Section}{Sections}
\Crefname{table}{Table}{Tables}
\crefname{table}{Tab.}{Tabs.}

\def\paperID{6551} %
\def\confName{ICCV}
\def\confYear{2025}

\title{
\dataname{}: Single-instance Multiview Multimodal and Multisetup \\ 3D Anomaly Detection Benchmark 
\vspace{-0.5cm}
}

\author{ 
    Alex Costanzino\textsuperscript{1} \hspace{0.3cm}
    Pierluigi Zama Ramirez\textsuperscript{1} \hspace{0.3cm}
    Luigi Lella\textsuperscript{1} \hspace{0.3cm}
    Matteo Ragaglia\textsuperscript{2} \hspace{0.3cm}
    Alessandro Oliva\textsuperscript{2} \\
    Giuseppe Lisanti\textsuperscript{1} \hspace{0.3cm}
    Luigi Di Stefano\textsuperscript{1} \\
    \small \textsuperscript{1}CVLab, University of Bologna, Italy \quad \small \textsuperscript{2}SACMI Imola, Italy \\
    \normalsize\url{https://alex-costanzino.github.io/SiM3D/}
    \vspace{-0.5cm}
}

\begin{document}

\maketitle

\begin{abstract}
\label{sec:abstract}
    \noindent
    We propose \dataname{}, the first benchmark considering the integration of multiview and multimodal information for comprehensive 3D anomaly detection and segmentation (ADS), where the task is to produce a voxel-based Anomaly Volume. 
    Moreover, \dataname{} focuses on a scenario of high interest in manufacturing: single-instance anomaly detection, where only one object, either real or synthetic, is available for training.
    In this respect, \dataname{} stands out as the first ADS benchmark that addresses the challenge of generalising from synthetic training data to real test data.
    \dataname{} includes a novel multimodal multiview dataset acquired using top-tier industrial sensors and robots.
    The dataset features multiview high-resolution images (12 {\tt Mpx}) and point clouds ($\sim$7M points) for 333 instances of eight types of objects, alongside a CAD model for each type. We also provide manually annotated 3D segmentation GTs for anomalous test samples.
    To establish reference baselines for the proposed multiview 3D ADS task, we adapt prominent singleview methods and assess their performance using novel metrics that operate on Anomaly Volumes. 
\end{abstract}
    
\section{Introduction}
\label{sec:introduction}
    
\begin{figure}[ht]
    \centering
    \includegraphics[width=0.84\linewidth]{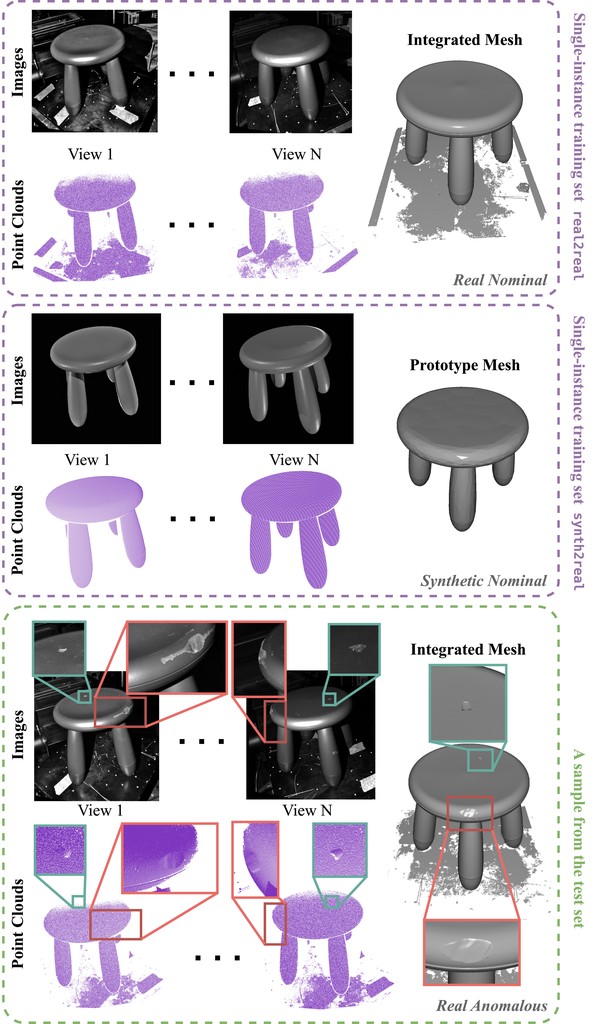}
    \caption{
        \textbf{\dataname{} dataset overview.} 
        From top to bottom: the single-instance real and synthetic training samples for object \emph{Plastic Stool}, one of the anomalous samples from the test set.
        }
    \label{fig:teaser}
\end{figure}

    In computer vision, anomaly detection and segmentation (ADS) involves identifying anomalous samples and localising their defects. 
    This task is challenging in industrial contexts due to the variability and rarity of defects. 
    Typically, ADS in the industry follows a \emph{cold-start} approach, where training is unsupervised and uses only defect-free (nominal) samples.
    Before the release of the MVTec AD dataset in 2019~\cite{bergmann2019mvtec}, only a limited number of ADS methods were proposed, with just 9 publications in major computer vision conferences (e.g., CVPR, ECCV, ICCV, WACV) over the five years before. 
    The introduction of MVTec AD marked a pivotal moment in the field, fostering the publication of roughly 172 papers in these venues over the next 5 years.
    This dataset not only stimulated the development of several ADS methods~\cite{patchcore2022roth, wang2023multimodal, cfm} but also promoted the creation of novel and more challenging benchmarks.
    For instance, MVTec 3D-AD and~\cite{bergmann2022mvtec} and Eyecandies~\cite{bonfiglioli2022eyecandies} introduced the first benchmarks deploying multimodal inputs to enhance ADS performance. Then, PAD~\cite{zhou2023pad} and Real-IAD~\cite{Wang_2024_CVPR_real_iad} introduced novel benchmarks that leverage multiple RGB views of objects. 
    Although these benchmarks tackle various settings, two real industrial issues remain unaddressed.

\begin{table*}[t]
\centering
    \resizebox{\linewidth}{!}{
        \begin{tabular}{lccccccccccc}
        \toprule
        \textbf{Dataset} & \textbf{Venue \& Year} & \textbf{2D} & \textbf{No. Pixels} & \textbf{3D} & \textbf{No. Points} & \textbf{Scenario} & \textbf{Train} & \textbf{Test} & \textbf{Setup} & \textbf{Task} & \textbf{No. Classes} \\ 
        \midrule
        MVTec AD        & CVPR 2019           & RGB Images  & 490k -- 1M            & ---        & ---               & Multi-instance   & Singleview    & Singleview   & \texttt{real2real}     & 2D & 15          \\ 
        MVTec 3D-AD     & VISAPP 2021         & RGB Images  & 160k -- 810k        & XYZ Images  & 10k -- 195k          & Multi-instance   & Singleview    & Singleview   & \texttt{real2real}     & 2D & 10         \\ 
        Eyecandies      & ACCV 2022           & RGB Images  & 262k               & Depths + Normals      & 262k              & Multi-instance   & Singleview    & Singleview   & \texttt{synth2synth}   & 2D & 10       \\ 
        MVTec LOCO AD   & IJCV 2022           & RGB Images  & 1.28M -- 2.05M      & ---        & ---               & Multi-instance   & Singleview    & Singleview   & \texttt{real2real}     & 2D & 5 \\ 
        VisA            & ECCV 2022           & RGB Images  & 1.5M               & ---        & ---               & Multi-instance   & Singleview    & Singleview   & \texttt{real2real}     & 2D & 12 \\ 
        PAD             & NeurIPS 2023        & RGB Images  & 12M                & ---        & ---               & Single-instance  & Multiview    & Singleview   & \texttt{real2real} + \texttt{synth2synth} & 2D & 20 \\ 
        Real3D-AD       & NeurIPS 2023        & ---         & ---                & Point Clouds & 43k -- 2.68M        & Multi-instance   & Multiview     & Singleview   & \texttt{real2real}          & 3D & 12 \\ 
        Real-IAD        & CVPR 2024           & RGB Images  & 10M                & ---        & ---               & Multi-instance   & Singleview    & Singleview   & \texttt{real2real}          & 2D & 30 \\ 
        \textbf{\dataname{}}            & \textbf{2025}                & \textbf{Grayscale Images} & \textbf{12M}          & \textbf{Point Clouds + Mesh} & \textbf{5M -- 7M} & \textbf{Single-instance}  & \textbf{Multiview}     & \textbf{Multiview}    & \textbf{\texttt{real2real} + \texttt{synth2real}} & \textbf{3D} & 8 \\ 
        
        \bottomrule
        \end{tabular}
        }
\caption{\textbf{Popular ADS benchmarks.} Chronologically sorted, the sequence hints at a growing interest in multimodal and multiview data.}
\label{tab:datasets}
\end{table*}

        \begin{figure}
        \centering
        \includegraphics[width=\linewidth]{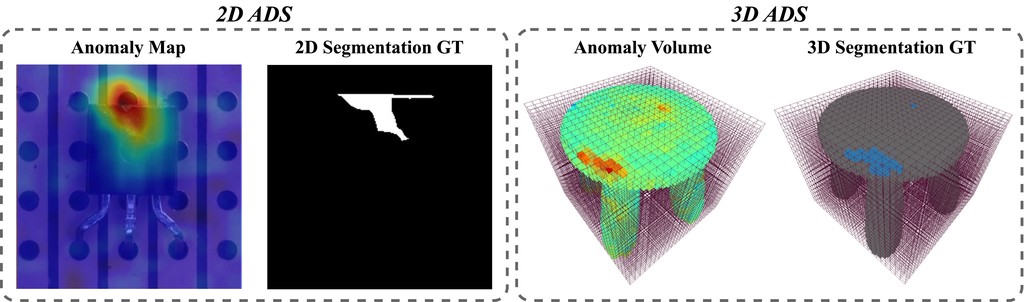}
        \caption{
        \textbf{2D vs. 3D anomaly detection and segmentation.}
        Typical ADS benchmarks rely on 2D Anomaly Maps and 2D ground-truths (left) to assess segmentation performance.
        \dataname{} proposes to do so by 3D Anomaly Volumes and 3D ground-truths (right).
        }
        \label{fig:2d_vs_3d_ADS}
    \end{figure}

    First, existing benchmarks have primarily addressed the problem of 2D ADS. 
    Consequently, current methods generate Anomaly Maps, i.e., images that encode the likelihood of each pixel belonging to a defect, as shown in~\cref{fig:2d_vs_3d_ADS} (left). 
    However, many industrial applications require precise localisation of defects in the 3D space to allow automatic intervention, reducing waste and optimising production.
    Benchmarks relying on multimodal input data, like those proposed in MVTec 3D-AD~\cite{bergmann2022mvtec} and Eyecandies~\cite{bonfiglioli2022eyecandies}, represent progress toward this goal by providing 3D information for each pixel within an image. 
    However, in these benchmarks, ADS methods are still expected to provide as output a 2D Anomaly Map, and objects are inspected from a single viewpoint.
    For comprehensive 3D anomaly detection, we must observe an object in its entirety, utilising captures from multiple viewpoints to detect defects across the entire 3D space. 
    This calls for methods capable of integrating multiview information to estimate the defect probability at each 3D location within an Anomaly Volume, as shown in~\cref{fig:2d_vs_3d_ADS} (right). Furthermore, as demonstrated by MVTec 3D-AD and Eyecandies, these methods should process multimodal information, i.e., RGB images as well as 3D information, to achieve more reliable anomaly detection.
    
    Secondly, current benchmarks typically provide multiple training samples per object class to capture the variability across nominal objects. 
    However, collecting many samples -- even nominal ones -- can be costly and challenging. 
    For example, line changeovers would demand acquiring new datasets and retraining models, both highly time-intensive processes. 
    Additionally, collecting large datasets increases the risk of errors, such as misjudged nominal samples, which may degrade model performance. 
    Moreover, manufactured objects typically closely resemble one another as they replicate the structure and materials of a synthetic CAD prototype. 
    Thus, the variability across nominal samples of manufactured objects is frequently minimal, with a single nominal item subsuming the information required to spot anomalies in other objects.  
    
    These considerations highlight the need for ADS techniques that may be effectively trained by a single nominal sample or even a synthetic prototype (e.g., an object’s CAD model) and then be able to generalise to other manufactured objects to identify anomalous ones effectively. 
    The ability to perform training by a single object instance, either real or synthetic, would avoid the need for extensive data collection, vastly facilitating the creation and adaptation of ADS models.  
    Yet, most current benchmarks focus on multiple training instances setups, and none of the prior works consider challenging synthetic (training) to real (test) setups. 
    
    To fill all the gaps highlighted by our considerations and stimulate interest in the development of multimodal and multiview 3D ADS methods, we propose the \underline{S}ingle-\underline{i}nstance \underline{M}ultiview \underline{M}ultimodal \underline{M}ultisetup dataset, dubbed \dataname{}.
    The dataset contains a series of high-resolution scans of manufactured objects, which can be nominal or anomalous.
    Each scan consists of multiple views, each featuring a greyscale image, its corresponding point cloud (i.e., multimodal data), and the ground-truth acquisition pose.
    In addition, an integrated 3D mesh is provided with each scan.
    For each object, a single nominal sample has been retained to be employed as a \emph{single-instance} training set, while the remaining samples, both nominal and anomalous, compose the test set.
    Moreover, to foster research on novel approaches that may address the domain gap between synthetic training data and real test samples in ADS, for each object of our dataset, we also provide a \emph{synthetic scan}, i.e., we deploy the available CAD model to obtain the same kind of data as in our high-resolution real scans, with images and point clouds rendered from the same poses. 
    An overview of the proposed dataset is depicted in~\cref{fig:teaser}.
    
    Based on our novel dataset, we create the first benchmark for 3D anomaly detection, where the task is to predict anomaly scores within a voxelized 3D Anomaly Volume rather than a 2D Anomaly Map (\cref{fig:2d_vs_3d_ADS}). 
    Peculiarly, our benchmark includes two setups based on the domain of the data available for training: in the \texttt{real2real} setup, a model is given access to a single nominal instance of an object (\cref{fig:teaser}, top), while in  \texttt{synth2real} the same kind of training data are obtained from the CAD model of the object (\cref{fig:teaser}, centre). 
    We adapt prominent ADS methods -- both unimodal and multimodal -- for the proposed task and evaluate them on our benchmark. 
    To assess segmentation performance, we provide accurate 3D ground-truths (\cref{fig:2d_vs_3d_ADS}) and extend ADS metrics to operate on anomaly scores and ground-truths specified on 3D voxel grids.
    Our experimental findings highlight the open challenges in the field and offer insights into possible future directions.

\section{Related Work}
\label{sec:related}
    The MVTec AD~\cite{bergmann2019mvtec} dataset significantly boosted anomaly detection research, inspiring new techniques and benchmarks for various challenges.
    MVTec AD provides a training set containing only nominal images and a test set that includes nominal and anomalous ones for several object categories.
    This unsupervised setup exemplifies many real-world applications in which anomalous objects are challenging to collect, and thus, it has also been adopted by more recent ADS benchmarks.
    Among them, MVTec LOCO~\cite{mvtec_loco} presents a challenging scenario with both structural anomalies (e.g., dents, holes) and logical anomalies (e.g., incorrect object combination).
    VisA~\cite{visa} introduced high-resolution images of complex scenes containing multiple instances of the same object.
    However, certain defects may be visible only by analysing objects' 3D geometry. 
    For this reason, multimodal datasets, such as MVTec 3D-AD~\cite{bergmann2022mvtec} and Eyecandies~\cite{bonfiglioli2022eyecandies} include 3D data alongside images. 
    For each sample in the dataset, MVTec 3D-AD provides images and pixel-aligned XYZ information captured by a 3D sensor, while Eyecandies contains synthetic images with pixel-aligned depths and normals.
    Recently, PAD~\cite{zhou2023pad} addressed pose-agnostic ADS with the first multiview dataset (MAD), featuring images of objects from various viewpoints. 
    In PAD, training samples are multiview RGB images with known poses, while test samples are singleview RGB images with unknown poses. MAD includes both synthetic and real objects, but its benchmarks do not explore synthetic-to-real generalisation.
    Real-IAD~\cite{Wang_2024_CVPR_real_iad} introduced a large-scale multiview dataset with RGB images covering various defect sizes. 
    Eventually, Real3D-AD~\cite{liu2023real3d} is the first point cloud anomaly detection benchmark where the task is to predict an anomaly score for each point. It uses front-and-back scans of nominal objects for training and a singleview point cloud for testing. 
    
    As shown in~\cref{tab:datasets}, the proposed \dataname{} benchmark offers several outstanding features, including very-high-resolution images (12 {\tt Mpx}), large point clouds/meshes containing 5-7 million elements captured with high-precision 3D sensors. 
    Moreover, it is designed for a highly relevant manufacturing scenario: single-instance anomaly detection, where only one object is available for model training. 
    To promote research toward even more cost-effective scenarios, ours is also the only benchmark that addresses the challenge of generalising from synthetic training data to real test data. Finally, \dataname{} stands out as the first benchmark that considers integrating multiview information at test time for comprehensive 3D anomaly detection, where the task consists of outputting a voxel-based Anomaly Volume.

    \noindent
    \textbf{ADS Methods.}
    Following the setup defined in the MVTec-AD benchmark, most ADS methods~\cite{liu2023deep} are designed to estimate 2D anomaly maps from a single image \cite{bergmann2018improving,zavrtanik2021draem,hou2021divide,ristea2022self, pirnay2022inpainting, Costanzino_2024_CVPR, wyatt2022anoddpm, reiss2021panda,reiss2021panda, yi2020patch, zhang2021anomaly, sohn2021learning, yoa2021self, li2021cutpaste, yang2023memseg, massoli2021mocca, yu2021fastflow, rudolph2021same, rippel2021modeling, gudovskiy2022cflow, chiu2023self, defard2021padim,Cohen2020SubImageAD, bergman2020deep}.
    One of the most popular methods is PatchCore~\cite{patchcore2022roth}, which uses feature extractors pre-trained on large datasets~\cite{caron2021emerging, oquab2023dinov2, he2022masked} to gather features from nominal samples into a memory bank. At test time, input sample features are compared to this bank to detect anomalies.
    Another more recent yet popular approach, EfficientAD~\cite{efficient_ad}, uses a lightweight feature extractor and a student-teacher model to identify anomalies based on feature prediction discrepancy. 
    The introduction of the MVTec 3D-AD benchmark has led to the development of several multimodal approaches that use images and 3D data to generate 2D anomaly maps \cite{horwitz2023back,wang2023multimodal,cfm,RudWeh2023}.
    Among them, BTF~\cite{horwitz2023back} is a memory bank approach inspired by PatchCore, in which each element is the concatenation of 2D and 3D features extracted by a pre-trained image backbone and a traditional point cloud descriptor, i.e., FPFH~\cite{fpfh}, respectively.
    M3DM~\cite{wang2023multimodal} improves BTF using powerful transformer-based pre-trained backbones, such as DINO-v1\cite{caron2021emerging} and Point-MAE~\cite{pang2022masked}, to extract features from RGB images and point clouds. It also learns to fuse 2D and 3D features into multimodal features.
    During inference, a trained One-Class SVM learns to aggregate modality-specific scores.
    AST~\cite{RudWeh2023} follows a Teacher-Student paradigm, in which the Teacher leverages Normalizing Flows to model the multimodal features distribution, while the Student employs a feed-forward network to model such distribution.
    This asymmetry in the architectures exacerbates the discrepancies between nominal and anomalous features at inference time.
    Recently, starting from the same transformer-based backbones as in~\cite{wang2023multimodal}, CFM~\cite{cfm} learns to map features from one modality to the other by nominal samples and then, at inference time, detects anomalies by pinpointing inconsistencies between observed and predicted features.
    In this paper, we employ the above-mentioned methods to investigate the challenges of \dataname{} by adapting them to work in a multiview setup.

\begin{figure*}[t]
    \centering
    \includegraphics[width=\linewidth]{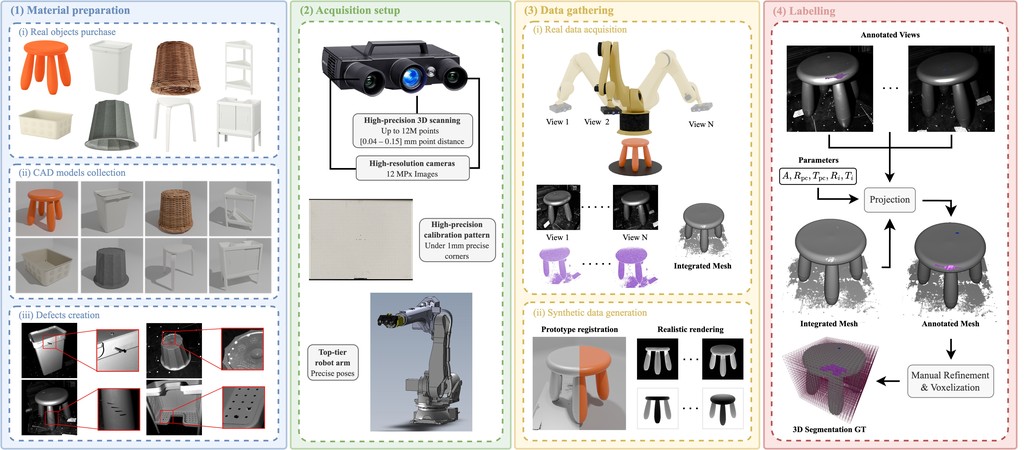}
    \caption{
    \textbf{\dataname{} dataset creation pipeline.}
    (1) We purchased eight types of objects with different properties, each associated with a CAD model. We altered part of the samples to create realistic anomalies;
    (2) We designed and calibrated an acquisition setup consisting of a high-precision 3D sensor with high-resolution cameras on top an industrial robot arm;
    (3) We acquired 360-degree scans of each object instance to populate the real split of the training and test sets, and rendered images and point clouds from the CAD models to create the synthetic split of the training sets;   
    (4) We created voxel-based segmentation GTs by a mixed 2D-3D labelling pipeline. 
    }
    \label{fig:dataset_creation}
\end{figure*}

\section{Dataset Creation}
\label{sec:dataset_creation}
        \noindent
        \textbf{Material preparation.}
        We scanned eight types of manufactured objects spanning different sizes, shapes, textures and materials, as shown in~\cref{fig:dataset_creation} (1.i). 
        For each object type, between 20 and 100 instances were purchased from a store. 
        We selected objects suitable for single-instance scenarios -- instances are manufactured according to a unique prototype model so that all nominal ones should be identical  -- and for which an official CAD prototype was available from the manufacturer. 
        The prototype CAD models are depicted in~\cref{fig:dataset_creation} (1.ii).
        To simulate various anomalies, we manually introduced different types of defects that represent realistic manufacturing flaws, as illustrated in~\cref{fig:dataset_creation} (1.iii)). 
        We altered the objects' appearance (e.g., paint modifications, \cref{fig:dataset_creation} (1.iii) -- bottom left), geometry (e.g., dents, \cref{fig:dataset_creation} (1.iii) -- bottom right), or both (e.g., contaminations, \cref{fig:dataset_creation} (1.iii) -- top left). 
        It is important to point out that only half of the instances within each object type are modified, while the remaining half are kept in their original state to serve as nominal data. 
        For each object type, one nominal instance will be used as the training set in the \texttt{real2real} experimental setup, while all other ones, both nominal and modified (i.e., anomalous), will make up the test set. 
        This very same test set is also used in the  \texttt{synth2real} setup, though the training data comes from CAD prototypes. 
    
        \noindent
        \textbf{Sensor.}
            We acquired our dataset with a top-tier 3D scanner, the Atos Q from ZEISS (\cref{fig:dataset_creation} (2.i)). It consists of a stereo pair with two 12 {\tt Mpx} greyscale cameras and a light projector. We consider the left camera to be our reference. The images have a resolution of $4096 \times 3000$ pixels. The point clouds can feature up to 12 million elements, with a point-to-point distance between 0.04 and 0.15 mm. The sensor is mounted on an industrial robot arm with high repeatability and precision (\cref{fig:dataset_creation} (2.iii)).
        
        \noindent
        \textbf{Calibration.}
            We calibrate the greyscale cameras with a standard chequerboard pattern, estimating the intrinsic parameters matrix, $A$, and the radial and tangential distortion coefficients, $\delta = [k_1, k_2, k_3, p_1, p_2]$. We also calibrate the rotation, $R_{pc}$, and translation, $T_{pc}$, from the point cloud to the left camera reference frame, with a precise dotted calibration pattern built by ZEISS (\cref{fig:dataset_creation} (2.ii)) that provides centre coordinates with precision $<1$ mm. More details are in the supplementary.
            
        \noindent
        \textbf{Real data acquisition.}
            The acquisition process consists of 360-degree scanning operations around all object instances of our dataset (see~\cref{fig:dataset_creation} (3.i)).
            We define specific viewpoints to capture data from multiple angles around the object. 
            These trajectories are sampled from concentric hemispheres that depend on the object's size. 
            We provide poses relative to a reference view, i.e., $v_1$. Thus, the pose for the view $i$, $v_i$, consists of a rotation $R_i$ and a translation $T_i$ that aligns $v_i$ to $v_1$, with $R_1$ the identity matrix and $T_1$ null. 
            Since in real industrial setups object instances would be acquired from as similar as possible viewpoints, we place markers in the scene to facilitate consistent positioning of objects of the same type. 
            Thus, thanks to the high repeatability of the robot arm, images, $I_i$, and point clouds, $P_i$, from the same viewpoint $v_i$ turn out similar across acquisitions of different instances of the same object type. 
            However, since the items were positioned manually with the help of the markers, slight displacements cannot be avoided. 
            This introduces some variability across the scans of the instances of an object, as it also typically occurs in real industrial pipelines. 
            For each object, after collecting data from all viewpoints, the ATOS Q sensor also returns an integrated mesh, $M$, obtained by registering all point clouds from the different views using a proprietary algorithm. 
            We collect and release the mesh to foster the development of methodologies that may also exploit this 3D representation. 
            The 3D reference frame is aligned to the one of the first point cloud $P_1$ taken from the first viewpoint $v_1$.

        \noindent
        \textbf{Synthetic data generation.}
            We employ the Blender \cite{blender} Python API to render synthetic data that simulates real acquisitions. 
            To ensure the synthetic and real data are aligned, we select a reference mesh from the collected real instances for each object type, load it in Blender, and align it to the prototype CAD model, as shown in (\cref{fig:dataset_creation} (3.ii) - left). 
            Then, we set the Blender camera parameters to the intrinsics, $A$, of the left camera of our Atos Q sensor and render images from the same viewpoints $v_i$ employed during the real acquisitions. 
            The RGB renders are converted into greyscale images to match the real images provided by our acquisition setup (\cref{fig:dataset_creation} (3.ii) - right). 
            Renderings have been obtained by exploiting Cycles Renderer, a path-tracing renderer able to provide realistic results with reflections, emissive surfaces, and physically-based lighting. 
            We also render depth maps (\cref{fig:dataset_creation} (3.ii) - right), which we project into 3D point clouds (see \cref{fig:teaser}, centre) using the intrinsic parameters, $A$.
    
        \noindent
        \textbf{Labelling.}
        \label{sec:labelling}
            Since some anomalies, such as paintings and minor scratches, are visible only in greyscale images, while others, such as dents, are better visible in 3D, we must consider both sensing modalities to obtain precise 3D segmentation ground-truths for anomalous samples. 
            Therefore, we develop a two-step labelling strategy, illustrated in~\cref{fig:dataset_creation} (4), that operates on both 2D images and 3D data. 
            First, we manually annotate all the view images, $I_i$, where anomalies are clearly visible, obtaining dense 2D segmentation masks. We employ a different ID and colour for every defect to ease the voxel AUPRO computation (see~\cref{sec:benchmark}). 
            Subsequently, for each annotated image, $I_i$, we leverage the associated integrated mesh, $M$, the intrinsic parameters $A$, the point cloud to camera reference frame transformation, $(R_\text{pc}, T_\text{pc})$, and the transformation between $v_i$ and $v_1$, $(R_i, T_i)$, to project the 2D annotations onto the integrated mesh. 
            Then, since the projection may not be precise due to occlusions and slanted surfaces, each ground-truth mesh is thoroughly inspected and manually refined using a 3D visualisation and editing software, i.e., CloudCompare \cite{cloudcompare}.
            The labelling process was conducted by a team of 4 experts, with a final cross-check of the results.
            Finally, the annotated mesh is converted into a voxel grid with a voxel size of 2 mm. 
            The choice of the size is determined by the smallest defect size discernible in the meshes. 
            In the conversion process, a voxel is labelled as anomalous if it intersects at least one triangle labelled as such in the annotated mesh.   

\begin{table}
    \centering
    \resizebox{\linewidth}{!}{%
        \begin{tabular}{l ccccccc} 

        \toprule
        
        \multirow{2}{*}{\textbf{Type}}     & Dimensions & Dimensions     & Total & Train Instances & \multicolumn{2}{c}{Test Instances} & Views\\
        
        & [cm] & [voxels] & Instances & Nominal & Nominal & Anomalous & per Instance   \\ 

        \midrule
        
        Plastic Stool      & $35 \times 35 \times 30$       & $330 \times 326 \times 317$ & 22          & 1 \texttt{real} or 1 \texttt{synth}    & 10            & 10 & 12  \\
        Rubbish Bin        & $26 \times 21 \times 33$       & $333 \times 339 \times 326$ & 42          & 1 \texttt{real} or 1 \texttt{synth}    & 20            & 20 & 12  \\
        Wicker Vase        & $17 \times 17 \times 15$       & $329 \times 309 \times 290$ & 22          & 1 \texttt{real} or 1 \texttt{synth}    & 10            & 10 & 12  \\
        Bathroom Furniture & $33 \times 33 \times 50$       & $371 \times 392 \times 363$ & 20          & 1 \texttt{real} or 1 \texttt{synth}    & 8             & 10 & 36  \\
        Container          & $20 \times 25 \times 10$       & $327 \times 308 \times 286$ & 94          & 1 \texttt{real} or 1 \texttt{synth}    & 46            & 46 & 12  \\
        Plastic Vase       & $12 \times 12 \times 9 $       & $326 \times 308 \times 286$ & 99          & 1 \texttt{real} or 1 \texttt{synth}    & 48            & 49 & 12  \\
        Wooden Stool       & $48 \times 42 \times 45$       & $331 \times 367 \times 327$ & 15          & 1 \texttt{real} or 1 \texttt{synth}    & 6             & 7  & 12  \\
        Sink Cabinet       & $44 \times 25 \times 50$       & $342 \times 349 \times 357$ & 19          & 1 \texttt{real} or 1 \texttt{synth}    & 9             & 8  & 36  \\

        \bottomrule
        
        \end{tabular}
    }
    \caption{\textbf{Dataset statistics.}} 
    \label{tab:stats}
\end{table}

\section{\dataname{} Benchmark}
\label{sec:benchmark}
    
    \subsection{Dataset, Tasks, and Metrics}
    \noindent
    \textbf{Dataset and Setups.}
        Using the procedure described in~\cref{sec:dataset_creation}, we collected 333 object instances belonging to the eight object types, as shown in~\cref{tab:stats}. We captured from 12 to 36 views for each instance, depending on the type. We split the data into training and test sets. For each type, the training set comprises a single instance, either real in the \texttt{real2real} setup or synthetic in the \texttt{synth2real} setup. The test set includes all other instances and is balanced between nominal and anomalous samples. The test set is the same for both setups.

    \noindent
    \textbf{Tasks.}
        Our benchmark focuses on the tasks of detecting and segmenting objects' anomalies using multiview and multimodal data. Specifically, the inputs to the methods consist of a set of images $\{I_i\}_{i=1}^n$ captured from different viewpoints alongside 3D data, $\mathbf{X}$, that capture an object entirely. The 3D data that a method can use are a set point clouds $\{P_i\}_{i=1}^n$ or depth maps $\{D_i\}_{i=1}^n$ taken from the same vantage points as the images, an integrated mesh $M$, or also a merged point cloud $P$.
        A method capable of multiview multimodal 3D anomaly \underline{detection} should take these inputs, integrate the 2D and 3D information from multiple views, and produce a global anomaly score for the test sample under inspection. On the other hand, a method pursuing multiview multimodal 3D anomaly \underline{segmentation} should yield a 3D output, where each 3D  position carries a score representing the likelihood of belonging to an anomaly. In our benchmark, a method for this task takes as input all images and 3D data, $(\{I_i\}_{i=1}^n, \mathbf{X})$ to output an Anomaly Volume, i.e. a voxel grid, $V \in \mathbb{R}^{X \times Y \times Z}$, where $X$, $Y$, and $Z$ are the grid dimensions and each voxel carries an anomaly score.
        For several reasons, we chose voxel grids as the 3D output representation of our benchmark. Firstly, as their grid structure resembles 2D Anomaly Maps, existing 2D segmentation methods and metrics can be easily extended. Moreover, voxel grids are suitable for representing every type of 3D defect, including missing parts.
        One major challenge with voxel grids pertains to choosing the voxel size. Ideally, we would want a size that allows accurate segmentation of even the smallest defects. 
        However, the memory occupancy of voxel grids scales cubically with the inverse of the voxel size. 
        Thus, we create ground-truth volumes using a voxel resolution of 2~mm, which enables precise localisation of even the smallest defects present in our dataset while keeping memory occupancy affordable. 

        \begin{figure}[t]
            \centering
            \includegraphics[width=\linewidth]{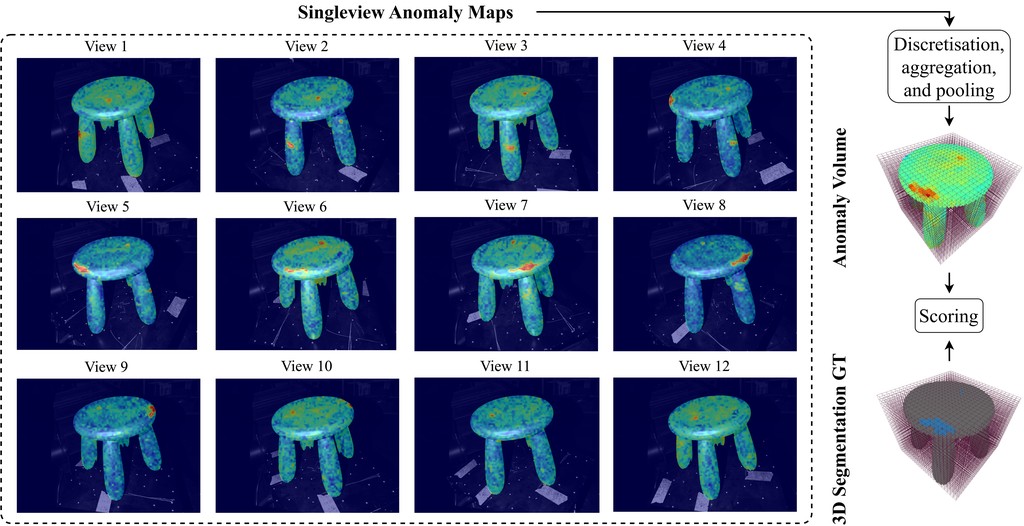}
            \caption{
            \textbf{Extending singleview methods.%
            } 
            Anomaly Maps obtained by processing singleview inputs with standard ADS methods are projected into the  3D and aggregated into a voxel grid, with each voxel containing a list of scores from multiple views. 
            The Anomaly Volume is obtained by voxel-wise max pooling. 
            }
            \label{fig:aggregation}
    \end{figure}

        \begin{figure*}
        \centering
        \includegraphics[width=0.98\linewidth]{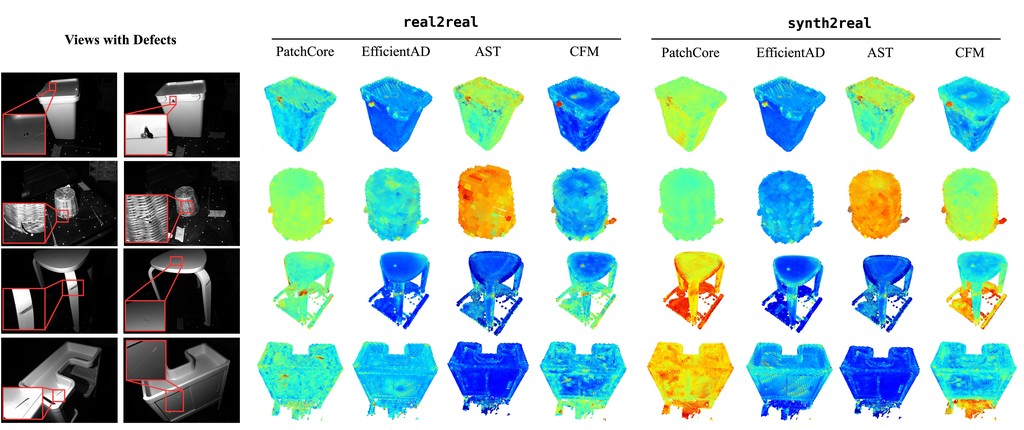}
        \caption{\textbf{Qualitatives.} Selected views with defects (left) and Anomaly Volumes for several methods (right) downsampled for visualisation.}
        \label{fig:qualitatives}
    \end{figure*}

    \begin{table*}[ht]
    \centering
    \resizebox{\linewidth}{!}{
        \begin{tabular}{ll ccccccccc ccccccccc}
        
            \cmidrule(lr){1-2} \cmidrule(lr){3-11} \cmidrule(lr){12-20}
            
            \multirow{2}[2]{*}{\textbf{{Method}}} & \multirow{2}[2]{*}{\textbf{{Modality}}} & \multicolumn{9}{c}{\texttt{real2real}} & \multicolumn{9}{c}{\texttt{synth2real}} \\

            \cmidrule(lr){3-11} \cmidrule(lr){12-20}
            
            &
            & 
            Pl. Stool & 
            Rub. Bin &
            W. Vase &
            B. Furn. &
            Cont. &
            Pl. Vase &
            W. Stool &
            Sink Cab. &
            \textbf{Mean} &
            Pl. Stool & 
            Rub. Bin &
            W. Vase &
            B. Furn. &
            Cont. &
            Pl. Vase &
            W. Stool &
            Sink Cab. &
            \textbf{Mean} \\
            
            \cmidrule(lr){1-2} \cmidrule(lr){3-11} \cmidrule(lr){12-20}
        
            PatchCore w/ WRN-101 & RGB & \underline{0.740} & \underline{0.987} & 0.636 & \underline{0.777} & \textbf{0.774} & 0.551 & \textbf{1.000} & 0.566 & \textbf{0.754} & 0.462 & 0.235 & 0.537 & 0.353 & \textbf{0.678} & 0.576 & 0.517 & 0.250 & 0.451 \\
            PatchCore w/ DINO-v2 & RGB & 0.500 & 0.958 & 0.636 & 0.622 & \underline{0.578} & 0.563 & \textbf{1.000} & 0.563 & 0.678 & \underline{0.958} & \textbf{0.897} & 0.413 & 0.464 & 0.207 & \textbf{0.684} & \underline{0.607} & 0.087 & \textbf{0.540} \\
            EfficientAD & RGB & 0.280 & 0.732 & 0.000 & \textbf{0.878} & 0.424 & \textbf{0.730} & \underline{0.928} & 0.712 & 0.585 & 0.123 & \underline{0.673} & 0.000 & \textbf{0.848} & 0.008 & 0.487 & \textbf{0.750} & \underline{0.662} & 0.443 \\
            AST & RGB & 0.537 & 0.566 & 0.611 & 0.515 & 0.573 & 0.496 & 0.839 & 0.537 & 0.584 & \textbf{1.000} & 0.344 & \textbf{1.000} & 0.525 & 0.482 & 0.388 & 0.428 & 0.187 & 0.544 \\

            \cmidrule(lr){1-2} \cmidrule(lr){3-11} \cmidrule(lr){12-20}
            
            BTF & RGB + Point Cloud & 0.421 & 0.217 & 0.504 & 0.565 & 0.545 & 0.471 & 0.678 & 0.424 & 0.478 & 0.462 & 0.294 & 0.520 & 0.444 & \underline{0.464} & 0.424 & 0.482 & 0.287 & 0.422 \\
            CFM~ w/ DINO-v2 + FPFH & RGB + Point Cloud & 0.198 & 0.301 & 0.074 & 0.515 & 0.483 & 0.456 & 0.732 & \underline{0.825} & 0.448 & 0.008 & 0.000 & 0.190 & 0.424 & 0.313 & 0.459 & 0.428 & 0.362 & 0.273 \\
            M3DM w/ DINO-v2 + FPFH & RGB + Point Cloud & 0.702 & \textbf{0.988} & \underline{0.661} & 0.545 & 0.556 & \underline{0.649} & 0.392 & 0.475 & 0.621 & 0.107 & 0.117 & \underline{0.735} & \underline{0.565} & 0.381 & \underline{0.586} & 0.214 & 0.512 & 0.402 \\
            AST & RGB + Depth & \textbf{0.950} & 0.927 & \textbf{0.785} & 0.474 & 0.542 & 0.470 & 0.428 & \textbf{0.925} & \underline{0.687} & 0.636 & 0.002 & 0.504 & 0.474 & 0.462 & 0.569 & 0.428 & \textbf{0.887} & \underline{0.495} \\

            \cmidrule(lr){1-2} \cmidrule(lr){3-11} \cmidrule(lr){12-20}

        \end{tabular}
    }
    \caption{\textbf{Anomaly detection results (I-AUROC).} Best results in \textbf{bold}, runner-ups \underline{underlined}.}
    \label{tab:detection_real_vs_synth}
\end{table*}

    \begin{table*}[ht]
    \centering
    \resizebox{\linewidth}{!}{
        \begin{tabular}{ll ccccccccc ccccccccc}
        
            \cmidrule(lr){1-2} \cmidrule(lr){3-11} \cmidrule(lr){12-20}
            
            \multirow{2}[2]{*}{\textbf{{Method}}} & \multirow{2}[2]{*}{\textbf{{Modality}}} & \multicolumn{9}{c}{\texttt{real2real}} & \multicolumn{9}{c}{\texttt{synth2real}} \\

            \cmidrule(lr){3-11} \cmidrule(lr){12-20}
            
            &
            & 
            Pl. Stool & 
            Rub. Bin &
            W. Vase &
            B. Furn. &
            Cont. &
            Pl. Vase &
            W. Stool &
            Sink Cab. &
            \textbf{Mean} &
            Pl. Stool & 
            Rub. Bin &
            W. Vase &
            B. Furn. &
            Cont. &
            Pl. Vase &
            W. Stool &
            Sink Cab. &
            \textbf{Mean} \\
            
            \cmidrule(lr){1-2} \cmidrule(lr){3-11} \cmidrule(lr){12-20}
                
            PatchCore w/ WRN-101 & RGB & 0.710 & 0.461 & 0.761 & 0.675 & 0.694 & 0.747 & 0.380 & 0.609 & 0.630 & \textbf{0.734} & 0.437 & 0.751 & 0.454 & 0.678 & 0.741 & 0.386 & 0.618 & 0.600 \\
            PatchCore w/ DINO-v2 & RGB & \underline{0.745} & 0.469 & 0.775 & 0.792 & \underline{0.709} & \underline{0.753} & 0.435 & 0.690 & 0.671 & 0.701 & 0.457 & 0.772 & 0.563 & 0.648 & 0.734 & 0.321 & 0.575 & 0.596 \\
            EfficientAD & RGB & 0.682 & 0.462 & 0.763 & 0.534 & 0.680 & 0.743 & 0.407 & 0.488 & 0.594 & 0.680 & 0.449 & 0.729 & 0.523 & 0.669 & \underline{0.747} & 0.431 & 0.361 & 0.573 \\
            AST & RGB & 0.720 & \underline{0.502} & \underline{0.789} & \underline{0.803} & \textbf{0.716} & \textbf{0.764} & \underline{0.446} & \underline{0.758} & \underline{0.687} & 0.726 & \underline{0.488} & \underline{0.791} & \textbf{0.806} & \underline{0.705} & \textbf{0.764} & \textbf{0.461} & \underline{0.695} & \underline{0.679} \\

            \cmidrule(lr){1-2} \cmidrule(lr){3-11} \cmidrule(lr){12-20}

            BTF & RGB + Point Cloud & 0.551 & 0.402 & 0.750 & 0.377 & 0.614 & 0.741 & 0.092 & 0.030 & 0.444 & 0.547 & 0.402 & 0.756 & 0.369 & 0.610 & 0.741 & 0.103 & 0.040 & 0.446 \\
            CFM~ w/ DINO-v2 + FPFH & RGB + Point Cloud & 0.597 & 0.415 & 0.768 & 0.505 & 0.640 & 0.743 & 0.315 & 0.321 & 0.538 & 0.523 & 0.413 & 0.753 & 0.542 & 0.621 & 0.741 & 0.222 & 0.314 & 0.516 \\ 
            M3DM w/ DINO-v2 + FPFH & RGB + Point Cloud & 0.733 & 0.452 & 0.767 & 0.702 & 0.702 & 0.752 & 0.288 & 0.126 & 0.565 & 0.690 & 0.454 & 0.770 & 0.461 & 0.645 & 0.734 & 0.318 & 0.132 & 0.526 \\
            AST & RGB + Depth & \textbf{0.750} & \textbf{0.503} & \textbf{0.792} & \textbf{0.807} & \textbf{0.716} & \textbf{0.764} & \textbf{0.467} & \textbf{0.798} & \textbf{0.699} & \textbf{0.751} & \textbf{0.502} & \textbf{0.793} & \underline{0.805} & \textbf{0.717} & \textbf{0.764} & \underline{0.450} & \textbf{0.785} & \textbf{0.695} \\
            
            \cmidrule(lr){1-2} \cmidrule(lr){3-11} \cmidrule(lr){12-20}
            
        \end{tabular}
    }
    \caption{\textbf{Anomaly segmentation results (V-AUPRO@1\%).} Best results in \textbf{bold}, runner-ups \underline{underlined}.}
    \label{tab:segmentation_real_vs_synth}
\end{table*}

    \begin{table}[ht]
    \centering
    \resizebox{\linewidth}{!}{
        \begin{tabular}{ll cc cc cc}
        
            \cmidrule(lr){1-4} \cmidrule(lr){5-6} \cmidrule(lr){7-8}
            
            \multirow{2}[2]{*}{\textbf{{Method}}} & \multirow{2}[2]{*}{{\textbf{Modality}}} & \multicolumn{2}{c}{{Features}} & \multicolumn{2}{c}{\texttt{real2real}} & \multicolumn{2}{c}{\texttt{synth2real}} \\

            \cmidrule(lr){3-4} \cmidrule(lr){5-6} \cmidrule(lr){7-8}

            &
            &
            2D &
            3D & 
            Detection &
            Segmentation &
            Detection &
            Segmentation \\
            
            \cmidrule(lr){1-4} \cmidrule(lr){5-6} \cmidrule(lr){7-8}
        
            \multirow{5}{*}{{PatchCore}} & RGB         & WRN-101 & -- & \textbf{0.754} & 0.630 & 0.451 & \textbf{0.600} \\
                                         & RGB         & DINO-v2 & -- & 0.678 & \textbf{0.671} & \textbf{0.540} & 0.596 \\                                        
                                         & Depth       & -- & WRN-101 & 0.582 & 0.486 & 0.507 & 0.484 \\
                                         & Depth       & -- & DINO-v2 & 0.600 & 0.496 & 0.246 & 0.471 \\
                                         & Point Cloud & -- & FPFH    & 0.415 & 0.464 & 0.313 & 0.460 \\ 
                
            \cmidrule(lr){1-4} \cmidrule(lr){5-6} \cmidrule(lr){7-8}

            \multirow{2}{*}{{BTF}}       & RGB + Point Cloud & WRN-101 & FPFH    & 0.478 & 0.444 & 0.422 & 0.446 \\
                                         & RGB + Depth       & WRN-101 & WRN-101 & \textbf{0.707} & \textbf{0.609} & \textbf{0.448} & \textbf{0.543} \\
            
            \cmidrule(lr){1-4} \cmidrule(lr){5-6} \cmidrule(lr){7-8}
            
            \multirow{2}{*}{{CFM}}       & RGB + Point Cloud & DINO-v2 & FPFH    & 0.448 & 0.538 & 0.273 & 0.516 \\
                                         & RGB + Depth & DINO-v2 & DINO-v2       & \textbf{0.548} & \textbf{0.663} & \textbf{0.311} & \textbf{0.620} \\

            \cmidrule(lr){1-4} \cmidrule(lr){5-6} \cmidrule(lr){7-8}
            \multirow{2}{*}{{M3DM}}      & RGB + Point Cloud & DINO-v2 & FPFH    & 0.621 & \textbf{0.565} & \textbf{0.402} & \textbf{0.526} \\
                                         & RGB + Depth       & DINO-v2 & DINO-v2 & \textbf{0.659} & 0.503 & 0.254 & 0.476 \\

            \cmidrule(lr){1-4} \cmidrule(lr){5-6} \cmidrule(lr){7-8}
            \multirow{2}{*}{{AST}}      & RGB & EffNet-B5 & --    & 0.584 & 0.687 & \textbf{0.544} & 0.679 \\
                                        & RGB + Depth & EffNet-B5 & EffNet-B5 & \textbf{0.687} & \textbf{0.699} & 0.495 & \textbf{0.695} \\

            \cmidrule(lr){1-4} \cmidrule(lr){5-6} \cmidrule(lr){7-8}
   
        \end{tabular}
    }
    \caption{\textbf{Anomaly detection and segmentation mean results with different input modalities.} Best results per method in \textbf{bold}.}
    \label{tab:detection_vs_segmentation}
\end{table}

    \noindent
    \textbf{Metrics.}
        Akin to standard practice in 2D ADS~\cite{bergmann2019mvtec}, we evaluate detection performance by the Area Under the Receiver Operator Curve (AUROC) of the global anomaly scores predicted by the methods. As we compute a global score for each object \textit{instance}, we name it instance-level AUROC (I-AUROC).
        Regarding segmentation performance, we evaluate methods with an adaptation of the popular Area Under the Per-Region Overlap (AUPRO) curve.
        First of all, we define the voxel-based Per-Region Overlap (PRO) metric as
            $\text{PRO}_t = \frac{1}{N} \sum_{n=1}^{N} \frac{\vert \overline{\Psi}_t \cap V_n \vert}{\vert V_n \vert}$
        , where $\overline{\Psi}_t$ is an Anomaly Volume binarised using a threshold, $t$, $N$ is the number of blobs, i.e., defects, in the ground-truth, and $V_n$ are the voxels belonging to that blob.
        Since during labelling, we annotated each defect in our dataset with a different ID (see \cref{sec:labelling}), the $N$ defect blobs in a ground-truth are easily identifiable.
        This metric is computed across multiple thresholds, $t$, estimated at different False Positive Rates (FPRs).
        We calculate the FPR using only non-empty voxels.
        We sample FPRs from 0 to the maximum value uniformly, we estimate the corresponding threshold $t$, and we binarise the score grid accordingly. In this way, we construct a curve that plots the PRO values against the corresponding FPRs.
        The segmentation performance is obtained by integrating such a curve up to a specific False Positive Rate. As proposed in~\cite{cfm}, we use 1\% as integration bound for the PRO curve to set a challenging bar representative of the requirements of real industrial applications.  Finally, we normalise the resulting area to the interval $[0, 1]$. We dub this metric voxel-level AUPRO (V-AUPRO).
        
    \subsection{Baselines}
        \noindent
        \textbf{Extending singleview methods to Multiview 3D ADS.}
            Current ADS  methods predict a 2D anomaly map from a singleview input, either unimodal, e.g., images, or multimodal, e.g., images and point clouds. To obtain a set of baselines for our novel multiview 3D ADS task, we extend popular and effective singleview methods by means of the 3D aggregation strategy illustrated in \cref{fig:aggregation}. Given the considered method, we process the data corresponding to an input view, $v_i$, so as to obtain a  2D Anomaly Map. Similarly to \cref{sec:labelling}, given the mesh, $M$, intrinsic parameters $A$, point cloud to camera reference frame transformation, $(R_\text{pc}, T_\text{pc})$ and view pose  $v_0$, $(R_i, T_i)$,  we can project each pixel of the 2D Anomaly Map to its corresponding 3D position. We discretise the 3D position, finding the corresponding voxel index. We associate that voxel with the corresponding anomaly score. We repeat the process for each view, keeping track of the scores that project onto the same voxels. Eventually, we compute the maximum among all scores projected into each voxel, obtaining the final Anomaly Volume. To extract the global score required by the I-AUROC, we compute the maximum across all the scores within the Anomaly Volume.

        \noindent   
        \textbf{Selected Methods.}
            We extend some of the most popular and effective unimodal and multimodal singleview ADS methods, starting from their official code, and evaluate them on \dataname{}.
            Regarding unimodal methods, we select PatchCore~\cite{patchcore2022roth},  with either WideResNet101~\cite{Zagoruyko2016WRN} or DINO-v2~\cite{oquab2023dinov2} as backbone, to extract features from images of all views, $\{I_i\}_{i=1}^N$ of the single-instance training set to create memory banks. 
            Moreover, we use EfficientAD~\cite{efficient_ad} using all views as their training set. 
            As for multimodal ADS methods, we use BTF~\cite{horwitz2023back} employing WideResNet101~\cite{Zagoruyko2016WRN} as the image encoder and FPFH~\cite{fpfh} as the algorithm to extract point cloud features. 
            We create memory banks using features extracted from images, $\{I_i\}_{i=1}^N$, and  point clouds, $\{P_i\}_{i=1}^N$, of all views of the single-instance training set.
            We use the same strategy with M3DM~\cite{wang2023multimodal}. As 2D feature extractor, we exploit DINO-v2. As point cloud feature extractor, instead of using a transformer backbone, i.e., Point-MAE~\cite{he2022masked}, we utilise FPFH to extract 3D features, similarly to BTF. Indeed, we would need a severe downsample (from $\sim$5 -- 7M to $\sim$8k points) to work with Point-MAE, and we verified that at such low resolution, most of the defects no longer show up in the point clouds. 
            Finally, we select  CFM~\cite{cfm}, using all views and clouds as a single training set. Based on the same consideration as for  M3DM, we use DINO-v2 and FPFH to produce 2D and 3D features.
            Due to computational constraints, all methods process images padded and downsampled to $1540 \times 1540$.  Moreover, for compatibility with the employed 2D feature extractors, which accept RGB images, we create 3-channel inputs by stacking our greyscale images thrice.  

        \noindent
        \textbf{3D Data pre-processing.}
            \label{subsec:preproc}
            To obtain less noisy data and improve the results of the above-mentioned baselines, we remove the background from 3D data. 
            We estimate a 3D plane using a RANSAC-based plane segmentation algorithm, which, iteratively, randomly selects a small subset of points (e.g., 10) to fit a candidate plane model. For the candidate planes, we compute the consensus set with threshold $\tau$ on the point-to-plane distance. After 1000 iterations, we return the best-fitting plane. To account for spurious background artefacts, we shift the plane by an offset $\alpha$ and finally remove all points below the shifted plane.
            Given the different sizes of the objects, for each object type, we select different optimal thresholds $\tau$ and offsets $\alpha$, as detailed in the supplementary material.
            We apply this background removal technique to the integrated mesh,  $M$,  due to the larger number of background points, which facilitates the RANSAC algorithm.
            Then, from the background-filtered integrated mesh,  we leverage ray-tracing techniques and known camera parameters to render, for each view, depth maps, $\{D_i\}_{i=1}^N$, pixel-aligned to the corresponding images, $\{I_i\}_{i=1}^N$ at the sensor resolution ($4096 \times 3000$).
            Finally, these depth maps are padded and downsampled to  $1540 \times 1540$, and the depth is lifted in 3D via the camera parameters to obtain dense and clean point clouds.
            The depth maps and point clouds obtained by rendering the pre-processed meshes serve as the input data for all the experiments presented in the subsequent section and will be available in the \dataname{} dataset.

\section{Benchmark Results}
\label{sec:experiments}

    \noindent
    \textbf{Main Results.}
        In ~\cref{tab:detection_real_vs_synth} and ~\cref{tab:segmentation_real_vs_synth}, we report the results obtained on the \dataname{} benchmark by extending the considered methods as Multiview 3D ADS baselines.
        By looking at the mean detection performance (\cref{tab:detection_real_vs_synth}), we notice how methods that natively process only RGB images outperform natively multimodal ones in both the \texttt{real2real} and \texttt{synth2real} setups. 
        With the notable exception of AST, this also stands for mean segmentation performance (\cref{tab:segmentation_real_vs_synth}).
        We posit that this is due to the main multimodal methods having been designed and tuned to process point clouds of the resolution featured by existing multimodal benchmarks, which is much lower than that of \dataname{}, i.e.  $\sim$ 8K vs $\sim$ 2.3M points (after downsampling).
        Indeed, AST is the only one designed to work natively with depth maps.
        Hence, our experiments suggest that current multimodal ADS methods do not scale to significantly higher resolutions.
        In turn, we could not even deploy the native 3D feature extractors of M3DM and CFM, the methods currently excelling on \cite{bergmann2022mvtec, bonfiglioli2022eyecandies}.
        Furthermore, we notice how methods relying on memory banks, such as PatchCore and M3DM, tend to attain slightly better mean detection and segmentation performance than their counterparts within the same modality that require substantial training, i.e., EfficientAD and CFM.
        We argue that this may be ascribed to the single-instance setup, which renders the latter category of approaches harder to optimise due to the limited number of training images, which are as many as the views.
        The methods most effectively extended to our multiview 3D task are Patchcore, which, equipped with either WRN-101 or DINO-v2, can achieve the best detection in both setups, and AST, achieving the best segmentation performance in both setups.
        Indeed, Patchcore yields I-AUROCs of 0.754 (WRN-101) and 0.540 (DINO-v2), while AST yields a V-AUPRO@1\%  of 0.699 and 0.695, in the \texttt{real2real} and \texttt{synth2real} setups, respectively.
        It is also worth pointing out how, due to the domain shift, we can notice a gap between the \texttt{real2real} and \texttt{synth2real} setups, especially in detection (0.754 vs. 0.540, I-AUROC) compared to segmentation performance (0.699 vs. 0.695, V-AUPRO@1\%).
        The qualitative results, shown in \cref{fig:qualitatives}, visually support the above observations.
        Overall, the relatively low, definitely far from saturated, performance figures provided by baselines designed by straightforward extension of state-of-the-art methods delivering impressive results in singleview 2D benchmarks hint at the novel and challenging nature of the task set forth by \dataname{}. Thus, our experimental findings by baseline approaches reveal the need for more advanced and specific methods and paradigms, both in the \texttt{real2real} and, even more so, the \texttt{synth2real} setups.
        
    \noindent
    \textbf{Alternative 3D Representations.}
        Given the lack of feature extractors amenable to processing high-resolution point clouds, we explore the use of depth maps
        to capture 3D cues. 
        The 2D grid structure of depth maps enables the deployment of popular backbones pre-trained on RGB data that support high-resolution inputs, such as DINO-v2. 
        Thus, we create and evaluate additional baselines using depth maps and report the results in ~\cref{tab:detection_vs_segmentation}.
        Similarly to greyscale images, we tile depth maps along the channel axis to ensure compatibility with the employed backbones.
        The results show that the use of depth maps rather than point clouds in multimodal methods tends to improve detection and segmentation performance by sizable margins, e.g., in the \texttt{real2real} setup, BTF yields 0.707 I-AUROC and 0.609 V-AUPRO when processing depths compared to 0.478 I-AUROC and 0.444 V-AUPRO when fed with point clouds (row 6 vs. row 7) while relying on the same WRN-101 image backbone. 
        Indeed, all multimodal methods but M3DM achieve their best performance in both tasks and setups by processing depth maps instead of point clouds.  
        Yet, as vouched by the results achieved by PatchCore, RGB images compare favourably to depth maps in terms of detection performance. 
        We ascribe this to the feature extractors used to compute depth features being pre-trained on RGB images, which may be conducive to yielding weaker features when fed with depth inputs. 
        This points out an intriguing research question: would it be possible to develop a foundation model to process depth maps effectively?

\section{Final Discussion}
\label{sec:discussion}
    We introduced \dataname{}, the first benchmark and dataset focused on integrating multiview and multimodal information for comprehensive 3D ADS.
    Our experiments highlight several open challenges. 
    First, a need for methods that can process multiview inputs more effectively, e.g., by taking all views as input rather than integrating singleview outputs, and that can be trained with a very limited number of nominal samples, e.g., a single instance.
    Second, training with synthetic data requires more robust techniques to address domain-shift challenges. %
    Finally, the current literature lacks 3D backbones capable of processing high-resolution point clouds. Employing depth maps seems a promising strategy to facilitate this processing, yet a foundational model for depth maps is still missing.
    We believe that \dataname{} may serve as a strong foundation for tackling these challenges and fostering future research toward 3D ADS.

    {
    \clearpage
    \section*{Acknowledgment}
        We acknowledge financial support from the National Recovery and Resilience Plan (NRRP), Mission 4, Component 2, Investment 1.1, Call No. 1409 published on 14 September 2022 by the Italian Ministry of University and Research (MUR), funded by the European Union.
        
        This work was also partially supported by the FSE+ 2021–2027 program under Article 24, paragraph 3, letter a) of Law 240/2010 and its subsequent amendments, as well as by Regional Resolution No. 693/2023 (Project Code: 2023-20090/RER), Grant CUP: J19J23000730002.
        
        We sincerely thank Lucia Gasperini for her contribution to dataset acquisition and labelling. We also extend our gratitude to Manuel Turrini for his support in setting up the acquisition process.
    }
    
    {
        \small
        \bibliographystyle{ieeenat_fullname}
        \bibliography{main}
    }

    {
        \clearpage
        \appendix
\renewcommand{\thesection}{S}

\setcounter{figure}{0}
\setcounter{table}{0}
\renewcommand{\thefigure}{\Alph{figure}}
\renewcommand{\thetable}{\Alph{table}}

\section*{Supplemental material}
    In this supplemental material, we provide additional qualitative results and insights on the design choices dealing with the creation of our dataset.
    \textbf{We would like to highlight that, in case of acceptance of the paper, \dataname{} will be publicly released along with the codebase created to implement the baselines and compute the performance evaluation metrics, so as to stimulate further research concerning the open challenges.}

    \subsection{Threshold employed for background removal}
        As anticipated in \cref{subsec:preproc} of the main paper, in \cref{tab:tau_and_alpha} we report the values of the threshold distance $\tau$ employed to determine the inliers of the consensus set and the offset $\alpha$ used to shift the fitted plane.
        The classes \emph{Wooden Stool} and \emph{Sink Cabinet} have not been filtered since the slight amount of background points does not provide enough support points to fit a plane with the employed algorithm.
        \begin{table}[ht]
    \centering
    \resizebox{0.5\linewidth}{!}{%
        \begin{tabular}{l  cc} 

        \toprule
        
        \textbf{Class}     & $\tau$ & $\alpha$ \\ 

        \midrule
        
        Plastic Stool      & 30 & -2 \\
        Rubbish Bin        & 30 & -2 \\
        Wicker Vase        & 60 & 10 \\
        Bathroom Furniture & 20 & 10 \\
        Container          & 30 & 2 \\
        Plastic Vase       & 60 & 2 \\
        Wooden Stool       & -- & -- \\
        Sink Cabinet       & -- & -- \\

        \bottomrule
        
        \end{tabular}
    }
    \caption{
    \textbf{Background removal $\tau$ and $\alpha$ values.}
    }
    \label{tab:tau_and_alpha}
\end{table}

    \subsection{Details on the defect distribution of the test set}
        In \cref{tab:defects} we report several statistics on the distribution of the defects on the test set, such as the number of test instances, the number of anomalies and the mean number of anomalies per defective instance.
        Even though each defect is inherently multimodal, they can be distinguished as:
        \emph{(i)} 3D, such as dents and bumps, characterised by significant structural variation but minimal changes in the visual appearance of the defective area;
        \emph{(ii)} 2D, such as scratches and marker strokes, in which we have a negligible structural variation and a significant deviation in the appearance of the defective area; 
        \emph{(iii)} multimodal, such as cracks and contaminations, in which both structural and appearance variations can be appreciated.
        \begin{table}
    \centering
    \resizebox{\linewidth}{!}{%
        \begin{tabular}{l  cc  cccc c} 

        \toprule
        
        \multirow{2}{*}{\textbf{Class}} & \multicolumn{2}{c}{Test Instances} & \multicolumn{4}{c}{No. Anomalies} & Mean Anomalies \\
        & Nominal & Anomalous & Tot. & 3D & 2D & Multimodal & per Instance \\

        \cmidrule(lr){1-1} \cmidrule(lr){2-3} \cmidrule(lr){4-7} \cmidrule(lr){8-8}
        
        Plastic Stool      & 10                  & 10                    & 32            & 11      & 18      & 3          & 3.2 \\
        Rubbish Bin        & 20                  & 20                    & 48            & 8       & 30      & 10         & 2.4 \\
        Wicker Vase        & 10                  & 10                    & 14            & 1       & 3       & 10         & 1.4 \\
        Bathroom Furniture & 8                   & 10                    & 32            & 0       & 24      & 8          & 3.2 \\
        Container          & 46                  & 46                    & 104           & 4       & 67      & 33         & 2.2 \\
        Plastic Vase       & 48                  & 49                    & 62            & 20      & 27      & 15         & 1.2 \\
        Wooden Stool       & 6                   & 7                     & 34            & 9       & 21      & 4          & 4.8 \\
        Sink Cabinet       & 9                   & 8                     & 27            & 4       & 15      & 8          & 3.3 \\

        \bottomrule
        
        \end{tabular}
    }
    \caption{\textbf{Test set defects statistics.}} 
    \label{tab:defects}
\end{table}

        Furthermore, we report in \cref{fig:distribution} the distribution of the size (in voxels) of the defects present in the test set of \dataname{}.
        The distribution highlights the predominance of smaller anomalies, which renders the benchmark particularly challenging.
        
\begin{figure}[ht]
    \centering
    \includegraphics[width=\linewidth]{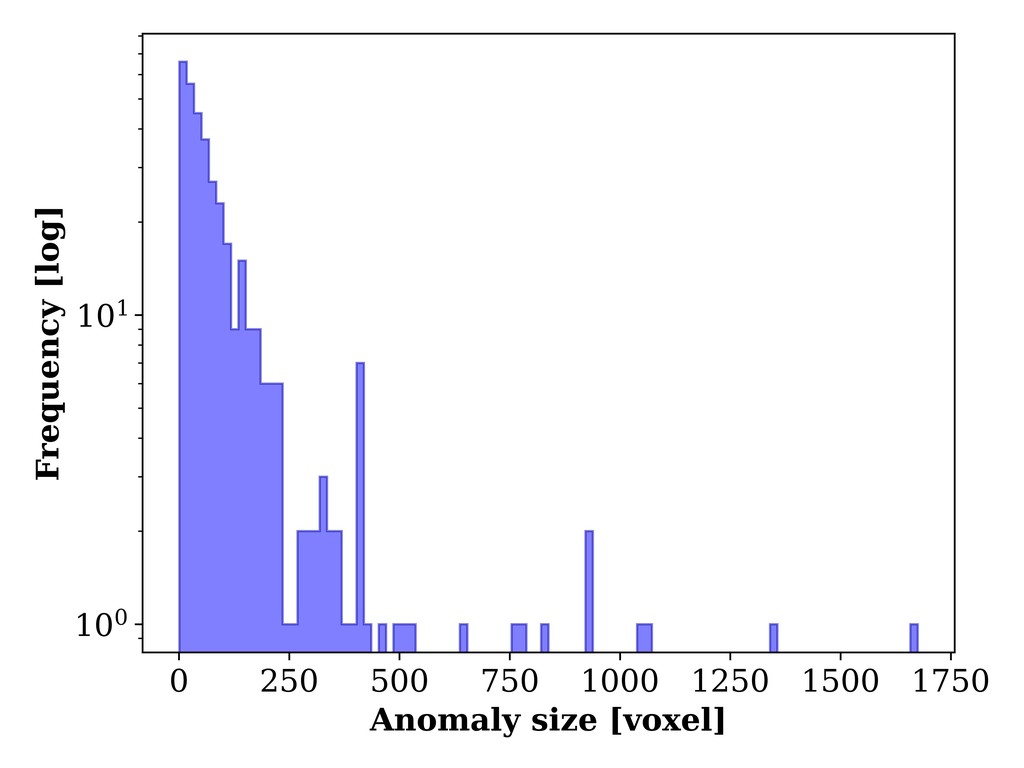}
    \caption{\textbf{Anomaly size distribution.} The proposed dataset contains predominantly smaller anomalies, making the overall benchmark particularly challenging.}
    \label{fig:distribution}
\end{figure}

    \subsection{Additional details on the calibration procedure}
        
\begin{figure*}[ht]
    \centering
    \includegraphics[width=\linewidth]{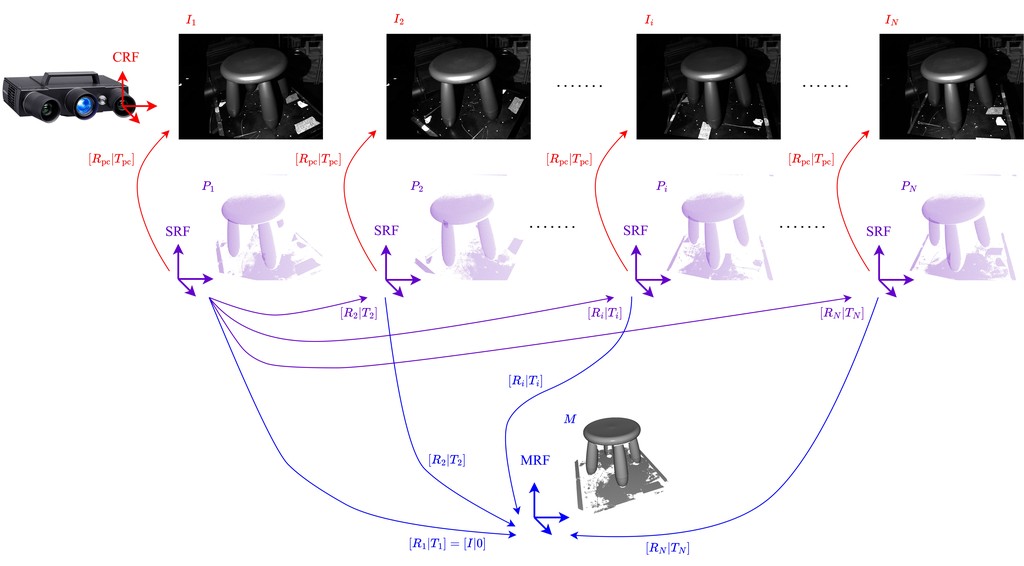}
    \caption{\textbf{Reference frames.}}
    \label{fig:reference_frames}
\end{figure*}

        The proposed pipeline deploys a custom procedure to estimate the transformation between the reference frame in which the Atos Q sensor provides the point cloud associated with a scan, from now on Sensor Reference Frame (SRF),  and the Camera Reference Frame of the left camera (CRF).  
        More in detail, we acquire several (i.e., 20) views, i.e., images and associated point clouds, of a high-precision dot pattern provided by ZEISS as part of the calibration toolkit of the Atos Q sensor (see \cref{fig:dataset_creation} (2) in the main paper). 
        Then, for each pair of images and point clouds, we apply the following steps:
        \begin{itemize}
            \item[(i)] Detect and refine the centres of the dots from the image by a classical circle detection algorithm;
            \item[(ii)] apply the Perspective-n-Points (PnP) algorithm between these 2D centres in the CRF and the known 3D coordinates of the centres expressed in the 3D reference frame attached to the calibration pattern to find the roto-translation between the CRF and the 3D reference frame attached to the calibration pattern;
            \item[(iii)] roto-translate the 3D centres of the dots -- expressed in the calibration pattern reference frame -- into the CRF, exploiting the transformation between the calibration pattern and the camera previously estimated by PnP;
            \item[(iv)] use the software tools  provided with the Atos Q to get the 3D coordinates of the centres of the dots into the SRF;
            \item[(v)] finally, given all the 3D-3D correspondences between the centres of the dots in the  SRF and the CRF obtained from the different views of the pattern,  we apply an absolute orientation algorithm (i.e., Kabsch-Umeyama) to estimate the roto-translation from the SRF to the CRF, i.e.  $[R_{\text{pc}} \vert T_{\text{pc}}]$.  
        \end{itemize}
            
        The accuracy of the estimation process can be assessed by computing the $\ell_2$ norm of the difference between the coordinates of centres brought from the SRF into the CRF by $[R_{\text{pc}} \vert T_{\text{pc}}]$ and those brought from the pattern reference frame into the CRF via the transformation obtained by PnP  
        This assessment is performed on an additional set of views with respect to those used to estimate $[R_{\text{pc}} \vert T_{\text{pc}}]$, resulting in a precision under 1 mm.

        In \cref{fig:reference_frames} we depict the different reference frames and the relations between them.
        We highlight that, with the Atos Q sensor,  the Mesh Reference Frame (MRF) is aligned in the 3D space to the  SRF associated with the first point cloud acquired while performing a 360-degree scanning of an object, hence $[R_{1} \vert T_{1}] = [I \vert 0]$.

\begin{figure*}[ht]
    \centering
    \includegraphics[width=\linewidth]{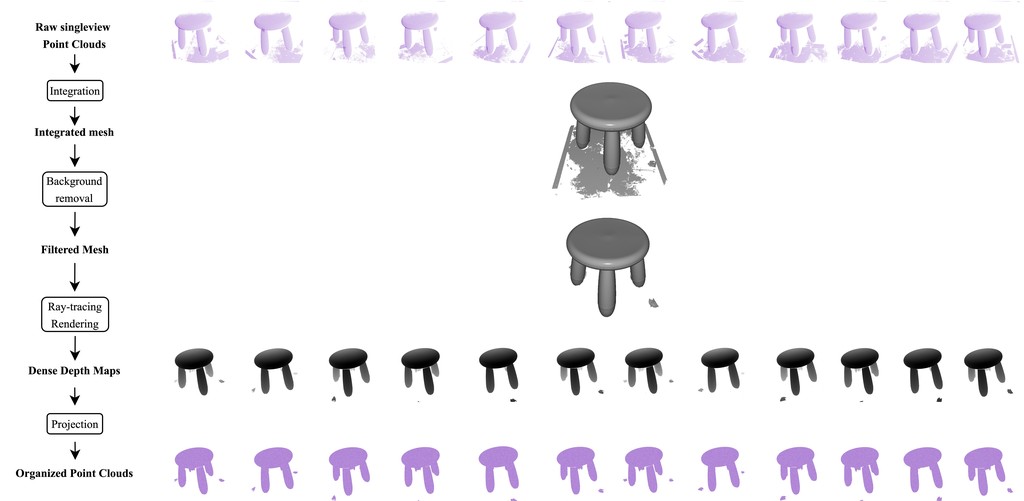}
    \caption{\textbf{3D Data pre-processing.}}
    \label{fig:3d_preprocess}
\end{figure*}

    \subsection{Additional details on the labelling procedure}
        As anticipated in \cref{sec:dataset_creation}, the labelling process involves transferring 2D annotations -- manually created on images -- to the 3D integrated mesh. 
        This is achieved by projecting the 3D mesh vertices into the 2D image space and associating each vertex with the ID colour (label information) carried by the corresponding pixel.
        Thus, the mesh vertices are first transformed by the inverse of the roto-translations $[R_{i} \vert T_{i}]$ between the mesh and the considered views. 
        Then, they are brought into  the CRF by employing $[R_{\text{pc}} \vert T_{\text{pc}}]$.
        Finally, the vertices' coordinates expressed in the CRF are projected onto the image plane of the 2D annotations by the intrinsic parameters of the camera, $A$.
        The corresponding pixel coordinates in the image are identified for each vertex, their ID colour (i.e. label) is extracted, and the information is assigned to the associated vertices of the 3D mesh, thereby lifting in 3D the annotations created on the original 2D inputs.
        Finally, the 3D mesh with vertex colours representing the annotations is manually refined.
        
    \subsection{Additional details on pre-processing of 3D data}
        In \cref{fig:3d_preprocess} we show the procedure to obtain the depth maps $\{D_i\}_{i=1}^n$, or organized point clouds $\{P_i\}_{i=1}^n$,  employed in the experiments described in \cref{sec:experiments}.
        In particular, after the raw singleview point clouds are acquired through a whole scan of the object, the ZEISS software shipped with the Atos Q integrates them to obtain a comprehensive mesh.
        Subsequently, we remove the background from this mesh with the procedure described in \cref{sec:benchmark}, obtaining a filtered mesh.
        Afterwards, by the knowledge of the roto-translation $[R_{\text{pc}} \vert T_{\text{pc}}]$ between the CRF and the SRF as well as those between the the roto-translations and the mesh, $[R_{i} \vert T_{i}]$ (see \cref{fig:reference_frames}), we can render the depth maps $\{D_i\}_{i=1}^n$ associated with each view by ray-tracing the cleaned mesh.
        Finally, to obtain clean and organized point clouds $\{P_i\}_{i=1}^n$, which are pixel-aligned to the corresponding greyscale images $\{I_i\}_{i=1}^n$, we can simply reproject the pixel coordinates in 3D via corresponding rendered depths and the inverse of the camera matrix, $A$.

\begin{figure*}[ht]
    \centering
    \includegraphics[width=0.99\linewidth]{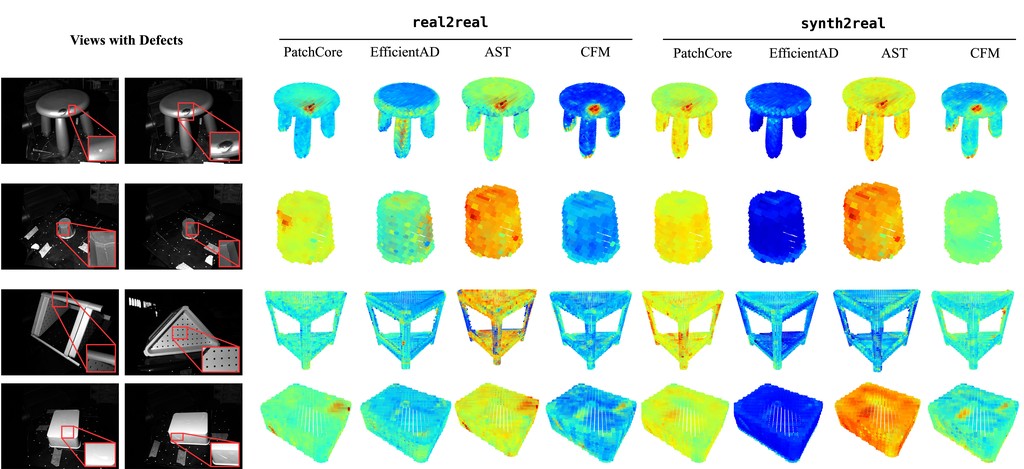}
    \caption{\textbf{Qualitatives.} Views with defects (left) and Anomaly Volumes for several methods (right) downsampled for visualisation.}
    \label{fig:qualitatives_supp}
\end{figure*}

\begin{figure*}[ht]
    \centering
    \includegraphics[width=0.99\linewidth]{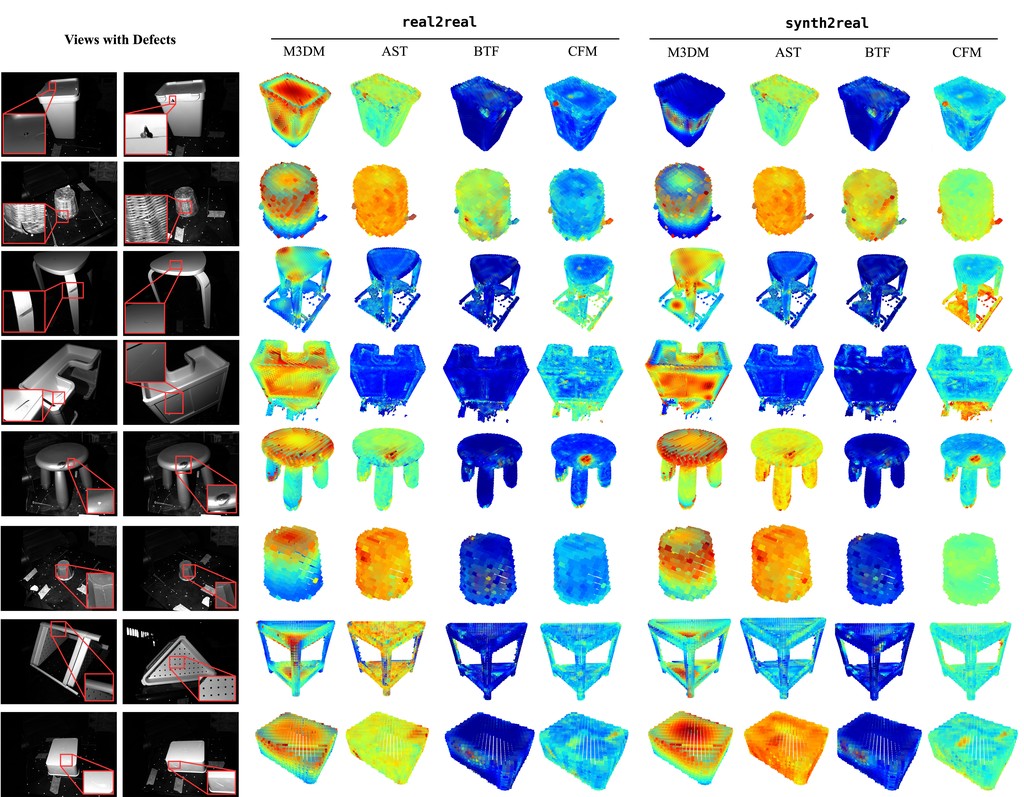}
    \caption{\textbf{Multimodal methods qualitatives.} Views with defects (left) and Anomaly Volumes for all multimodal methods (right) downsampled for visualisation.}
    \label{fig:all_multimodal_qual}
\end{figure*}

    \subsection{Additional details on the baselines}
        For the following baselines, we process greyscale images $\{I_i\}_{i=1}^n$, depth maps $\{D_i\}_{i=1}^n$ or organized point clouds $\{P_i\}_{i=1}^n$ downsampled to the common size of $1540 \times 1540 \times 3$, the highest achievable with the available hardware.
        
        Input data processed with WideResNet101, either greyscale images or depth maps, returns features from the second and third layers, yielding feature maps with dimensions equal to $28 \times 28 \times 1536$.
        Input data processed with DINO-v2, either greyscale images or depth maps,  return features from the last layer, yielding feature maps with dimensions equal to $110 \times 110 \times 768$.
        Input data processed with FPFH features, namely point clouds, are processed considering a voxel size equal to 2 mm in order to match the resolution attainable from the 3D voxel ground-truths and yield feature maps with dimensions equal to $1540 \times 1540 \times 33$.
        The spatial resolution of these maps is then downsampled to match either the spatial resolution of the WideResNet101 or DINO-v2 features, hence $28 \times 28$ or $110 \times 110$.
        
        Unlike the common practice, the 2D Anomaly Maps obtained by the methods are not Gaussian-blurred since the subsequent 3D aggregation and discretisation introduce smoothing.

    \noindent
    \textbf{PatchCore.}
        PatchCore~\cite{patchcore2022roth} is a singleview, image-based ADS method that employs WideResNet101 to extract features from training data, which are subsequently stored in a memory bank. 
        During inference, features from test samples are queried against this memory bank to compute an anomaly score.
        
        We adapted this method on \dataname{}, by either processing greyscale images or depth maps, using either its original feature extractor, WideResNet101, or the DINO-v2 backbone.
        We created the coreset with a 10\% coverage and 0.9 projection radius and selected 3 as a reweight parameter for the anomaly map computation.
        
    \noindent
    \textbf{EfficientAD.}
        Efficient~\cite{efficient_ad} is a singleview, image-based ADS method that employs a Teacher--Student paradigm based on patch description networks paired with an autoencoder pre-trained on WideResNet101.

        We implement EfficientAD by disabling the Teacher normalisation since there is no validation set available, and we upsample the intermediate encoder outputs to match the size of the Teacher and the Students fed with $1540 \times 1540 \times 3$ images.
        Moreover, we train the Students for 1000 epochs, unlike the 70000 expected from the adopted implementation, since the loss tends to stall earlier due to the limited number of training images that characterise our single-instance setup. 
        
    \noindent
    \textbf{Back to the Features.}
        BTF~\cite{horwitz2023back} is a singleview, multimodal ADS method that, similarly to PatchCore, employs WideResNet101 to extract 2D features from RGB images, and relies on FPFH to extract 3D features from point clouds. The 2D and 3D features are subsequently concatenated and stored in a single memory bank. 
        During inference, 2D and 3D features from test samples are also concatenated and queried against this memory bank to compute an anomaly score.
        
        We also implement BTF by using WideResNet101 to process both grayscale images and depth maps.
        We created the coreset with a 10\% coverage and 0.9 projection radius and selected 3 as a reweight parameter for the anomaly map computation.

    \noindent
    \textbf{Crossmodal Feature Mapping.}
        CFM~\cite{cfm} is a singleview, multimodal ADS method that exploits MLPs to map features from one modality to the other on nominal samples and then detect anomalies by pinpointing inconsistencies between observed and mapped features.
        This solution leverages DINO-v1 and Point-MAE to extract features from RGB images and point clouds, respectively.
    
        Since \dataname{} contains high-resolution images and point clouds, we adopt DINO-v2 and FPFH as feature extractors.
        Alternatively, to reduce computational demands, we also implement CFM by using DINO-v2 to process both greyscale images and depth maps.
        We trained the crossmodal feature mappings for 50 epochs, following the original implementation, with a unitary batch size to limit memory consumption.

    \noindent
    \textbf{Multi-3D-Memory.}
        M3DM~\cite{wang2023multimodal} is a singleview, multimodal ADS method that employs DINO-v1 and Point-MAE to extract features from RGB images and point clouds, which are subsequently stored in memory banks.
        Moreover, it also learns a function to fuse 2D and 3D features into multimodal features, which are then stored in memory banks alongside those computed from the individual modalities.
        During inference, features from test samples are queried against the memory banks to compute anomaly scores, which are then aggregated with One-Class SVMs.
        
        Given that \dataname{} contains high-resolution images and point clouds, we adopt DINO-v2 and FPFH as feature extractors.
        To reduce computational demands, we also implement M3DM by using DINO-v2 to process both greyscale images and the rendered depth maps.
        Furthermore, we disabled the feature fusion module due to computational limitations introduced by the high-resolution features.
        Following the original implementation, we created both coresets with a 10\% coverage and 0.9 projection radius and selected 1 as a reweight parameter for the image-based anomaly map and 0.1 as a reweight parameter for the point cloud-based anomaly map.
        Moreover, both One-Class SVMs are trained with a $\nu$ parameter equal to 0.5 and a maximum number of SGD iterations fixed to 1000.

    \noindent
    \textbf{Asymmetric Student-Teacher.}
        AST~\cite{RudWeh2023} is a singleview, multimodal ADS method that employs EfficientNet-B5 to extract features from RGB images and depth maps.  
        Such features are subsequently employed to optimise a Normalizing Flow as Teacher network and a feed-forward network as a Student network.
        The idea is that, after optimisation, both networks are able to reconstruct nominal samples, begetting low discrepancies, while failing to reconstruct anomalous samples.
        Since these two networks present different architectures, the way in which they fail to reconstruct anomalous samples is different, hence discrepancies will be exacerbated, highlighting anomalies.
        Due to the fact that this method work on the features, it is triviality extendable to multiple modalities.
        
        Given that \dataname{} contains high-resolution images and point clouds, we deployed AST by passing to the framework $1540 \times 1540 \times 3$ images and depth maps.
    
    \begin{table*}[ht]
        \centering
            \resizebox{\linewidth}{!}{
                \begin{tabular}{l ccccccccc}
                    \midrule
                    \textbf{} &
                    Pl. Stool & 
                    Rub. Bin &
                    W. Vase &
                    B. Furn. &
                    Cont. &
                    Pl. Vase &
                    W. Stool &
                    Sink Cab. &
                    \textbf{Mean} \\
                    \midrule
                
                    FID \texttt{real2test}  & 10.088 & 8.844 & 7.345 & 15.784 & 12.590 & 9.560 & 8.910 & 23.750 & 12.108 \\
                    FID \texttt{synth2test} & 48.744 & 44.565 & 57.302 & 40.900 & 51.900 & 55.194 & 24.040 & 34.060 & 44.588 \\
                    $\Delta$ FID   & 38.656 & 35.721 & 49.957 & 25.116 & 39.310 & 45.634 & 15.130 & 10.310 & 32.480 \\ 
                    $\Delta$ I-AUROC (PatchCore)  & 0.458 & 0.061 & 0.223 & 0.158 & 0.371 & 0.121 & 0.393 & 0.476 & 0.282 \\
                    $\Delta$ V-AUPRO (PatchCore) & 0.044 & 0.012 & 0.003 & 0.229 & 0.061 & 0.019 & 0.114 & 0.115 & 0.075\\
    
                    \midrule
                \end{tabular}
            }
        \caption{\textbf{Data quality assessment and impact on performance.} The average Fréchet Inception Distance between the train samples and the test samples for both \texttt{real2real} and \texttt{synth2real} setups is reported.}
        \label{tab:FID}
    \end{table*}

    \begin{figure}[ht]
        \centering
        \includegraphics[width=0.75\linewidth]{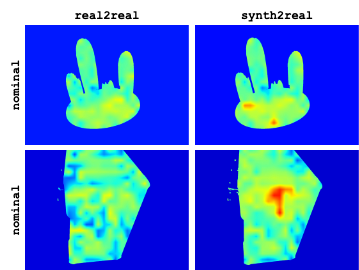}
        \caption{\textbf{Outliers.} The same test views inferred with the same model, trained either with \texttt{real2real} or \texttt{synth2real} train set, highlights the presence of outliers in the \texttt{synth2real} scenario.}
        \label{fig:outliers}
    \end{figure}
    
    \subsection{Assessment of data generation quality and insight of failure cases of the \texttt{synth2real} setup}
        We report in~\cref{tab:FID} the Fréchet Inception Distance (FID) between the train set, either from the \texttt{real2real} or \texttt{synth2real} setup, and the test set, along with the performance gaps ($\Delta$ I-AUROC, $\Delta$ V-AUPRO) for PatchCore.
        We do not observe any correlation between $\Delta$ FID and $\Delta$ I-AUROC or $\Delta$ V-AUPRO.
        Instead, in the \texttt{synth2real} scenario, we noticed strong outliers in anomaly maps of \textbf{nominal samples} (see \cref{fig:outliers}), for some objects. 
        These outliers have a strong impact on detection since they increase the false detection rate, while affecting the segmentation performance much less. 

    \subsection{Ancillary experiments to assess the \dataname{} challenges}
        As suggested during the peer-reviewing process, we ran the following experiments to gain more insight into the unique challenges set forth by \dataname{}:
        \begin{itemize}
            \item We treated the task as single-view ADS by deploying PatchCore, obtaining a mean I-AUROC of 0.488 vs. 0.754 of our multiview approach, \textbf{highlighting the necessity of addressing the task in a multiview fashion}. Notice that we cannot compare the segmentation performance since the ground-truths are voxel grids;
            \item We ran the official implementation of PatchCore on grayscale MVTec AD, obtaining a mean I-AUROC of 0.988 vs. 0.991, and a mean AUPRO@30\% of 0.929 vs. 0.935, both official results from \cite{patchcore2022roth}, \textbf{highlighting that benchmark complexity does not stem from grayscale images}.
        \end{itemize}
        As for metrics, I-AUROC is directly comparable to other benchmarks; hence, the \textbf{general lower performance confirms that SiM3D is more challenging}. Segmentation performance cannot be compared since the ground-truths of \dataname{} are voxel grids.

    \subsection{Assessment of baselines on 2D and multimodal anomalies}
        As suggested during the peer-reviewing process, we performed a comparative analysis of some baselines while disentangling the different kinds of anomalies.
        To this end, starting from \cref{tab:defects}, we split the anomalies into 2D and 3D+multimodal anomalies, hence, unifying 3D defects with the multimodal ones.
        Indeed, we argue that, while it is possible to unambiguously consider some anomalies, such as marker strokes and white-outs, as 2D-only, it is more difficult to pinpoint 3D-only anomalies, as structural deviations in the geometry, more often than not, tend to manifest themselves also in the image space.
        
        After that, we selected the best-performing algorithm working on images (PatchCore w/ DINO-v2) and multimodal data (AST), 
        The experiments, reported in \cref{tab:2d_vs_multimodal}, show that PatchCore, the image-based method, tend to perform best on 2D anomalies and worse on multimodal anomalies, obtaining an overall weaker performance when all the anomalies are considered.
        On the other hand, AST, the multimodal method, tend to perform well on both 2D and multimodal anomalies, obtaining an overall stronger performance when all the anomalies are considered.
        These findings further confirm the necessity of a multimodal analysis when working in the \dataname{} setup.
        \begin{table*}[ht]
    \centering
    \resizebox{\linewidth}{!}{
        \begin{tabular}{llc ccccccccc ccccccccc}
        
            \cmidrule(lr){1-3} \cmidrule(lr){4-12} \cmidrule(lr){13-21}
            
            \multirow{2}[2]{*}{\textbf{{Method}}} & \multirow{2}[2]{*}{\textbf{{Modality}}} & \multirow{2}[2]{*}{\textbf{{Anomalies}}} & \multicolumn{9}{c}{\textbf{Detection}} & \multicolumn{9}{c}{\textbf{Segmentation}} \\

            \cmidrule(lr){4-12} \cmidrule(lr){13-21}

            &
            &
            & 
            Pl. Stool & 
            Rub. Bin &
            W. Vase &
            B. Furn. &
            Cont. &
            Pl. Vase &
            W. Stool &
            Sink Cab. &
            \textbf{Mean} &
            Pl. Stool & 
            Rub. Bin &
            W. Vase &
            B. Furn. &
            Cont. &
            Pl. Vase &
            W. Stool &
            Sink Cab. &
            \textbf{Mean} \\
            
            \cmidrule(lr){1-3} \cmidrule(lr){4-12} \cmidrule(lr){13-21}
        
            PatchCore w/ DINO-v2 & RGB & All        & 0.500 & 0.958 & 0.636 & 0.622 & 0.578 & 0.563 & 1.000 & 0.563 & 0.678 & 0.745 & 0.469 & 0.775 & 0.792 & 0.709 & 0.753 & 0.435 & 0.690 & 0.671 \\
            PatchCore w/ DINO-v2 & RGB & 2D         & 0.646 & 0.902 & 0.602 & 0.723 & 0.678 & 0.498 & 0.980 & 0.612 & 0.705 & 0.832 & 0.498 & 0.674 & 0.801 & 0.718 & 0.687 & 0.455 & 0.678 & 0.667 \\
            PatchCore w/ DINO-v2 & RGB & Multimodal & 0.425 & 0.573 & 0.534 & 0.439 & 0.592 & 0.632 & 0.754 & 0.587 & 0.567 & 0.276 & 0.354 & 0.456 & 0.548 & 0.698 & 0.603 & 0.404 & 0.596 & 0.491 \\

            \cmidrule(lr){1-3} \cmidrule(lr){4-12} \cmidrule(lr){13-21}
            
            AST w/ EffNet-B5 & RGB + Depth & All        & 0.950 & 0.927 & 0.785 & 0.474 & 0.542 & 0.470 & 0.428 & 0.925 & 0.687 & 0.750 & 0.503 & 0.792 & 0.807 & 0.716 & 0.764 & 0.467 & 0.798 & 0.699 \\
            AST w/ EffNet-B5 & RGB + Depth & 2D         & 0.910 & 0.905 & 0.773 & 0.427 & 0.539 & 0.320 & 0.423 & 0.892 & 0.648 & 0.654 & 0.443 & 0.730 & 0.687 & 0.708 & 0.797 & 0.303 & 0.679 & 0.625 \\
            AST w/ EffNet-B5 & RGB + Depth & Multimodal & 0.945 & 0.895 & 0.698 & 0.410 & 0.498 & 0.543 & 0.413 & 0.904 & 0.663 & 0.738 & 0.475 & 0.765 & 0.789 & 0.893 & 0.683 & 0.564 & 0.723 & 0.703 \\
            
            \cmidrule(lr){1-3} \cmidrule(lr){4-12} \cmidrule(lr){13-21}

        \end{tabular}
        }
    \caption{\textbf{Anomaly detection and segmentation results considering different kinds of anomalies.}}
    \label{tab:2d_vs_multimodal}
\end{table*}

    \subsection{Attribution of existing assets}
        We adapted the baselines described in the main paper starting from the following codebases:        
        \begin{itemize}
            \item PatchCore: \url{https://github.com/eliahuhorwitz/3D-ADS} released under MIT License;
            \item EfficientAD: \url{https://github.com/nelson1425/EfficientAD} released under Apache License 2.0;
            \item BTF: \url{https://github.com/eliahuhorwitz/3D-ADS} released under MIT License;
            \item M3DM: \url{https://github.com/nomewang/M3DM} released under MIT License;
            \item CFM: \url{https://github.com/CVLAB-Unibo/crossmodal-feature-mapping} released under non-commercial use only license;
            \item AST: \url{https://github.com/marco-rudolph/AST} released under MIT License.
        \end{itemize}

\begin{figure*}[t]
    \centering
    \includegraphics[width=0.49\linewidth]{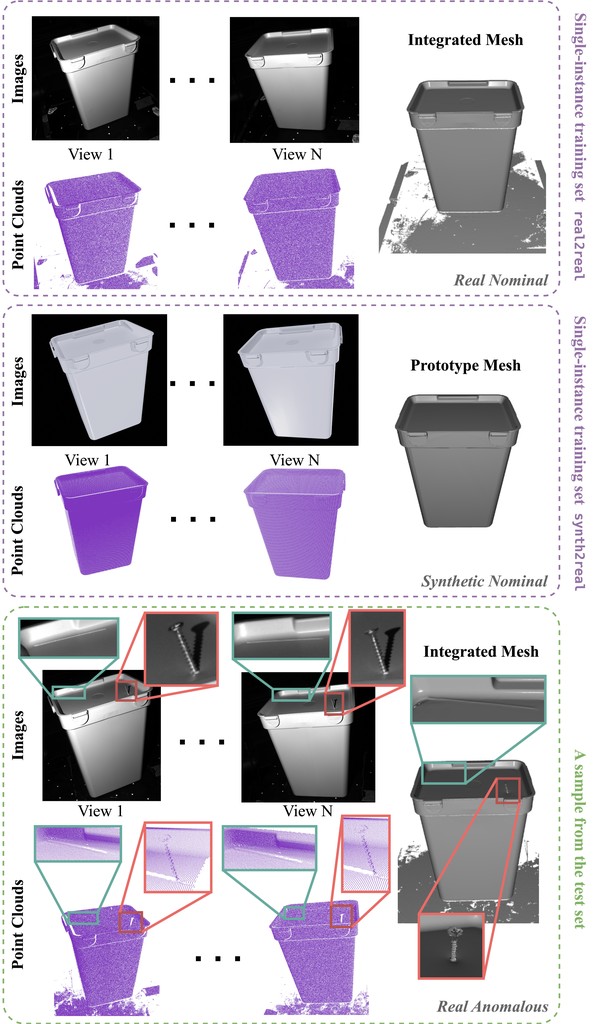}
    \includegraphics[width=0.49\linewidth]{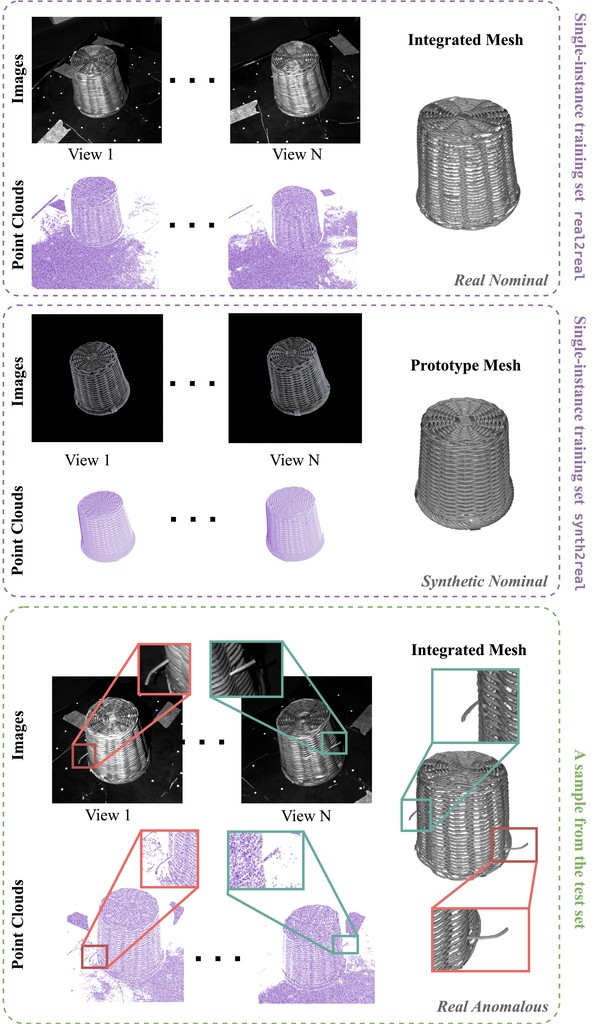}
    \caption{
        \textbf{\dataname{} dataset overview.} 
        From top to bottom: the single-instance real and synthetic training samples for objects \emph{Rubbish Bin} and \emph{Wicker Vase}, one of the anomalous samples from the test set.
        }
    \label{fig:overview_a}
\end{figure*}

\begin{figure*}[ht]
    \centering
    \includegraphics[width=0.49\linewidth]{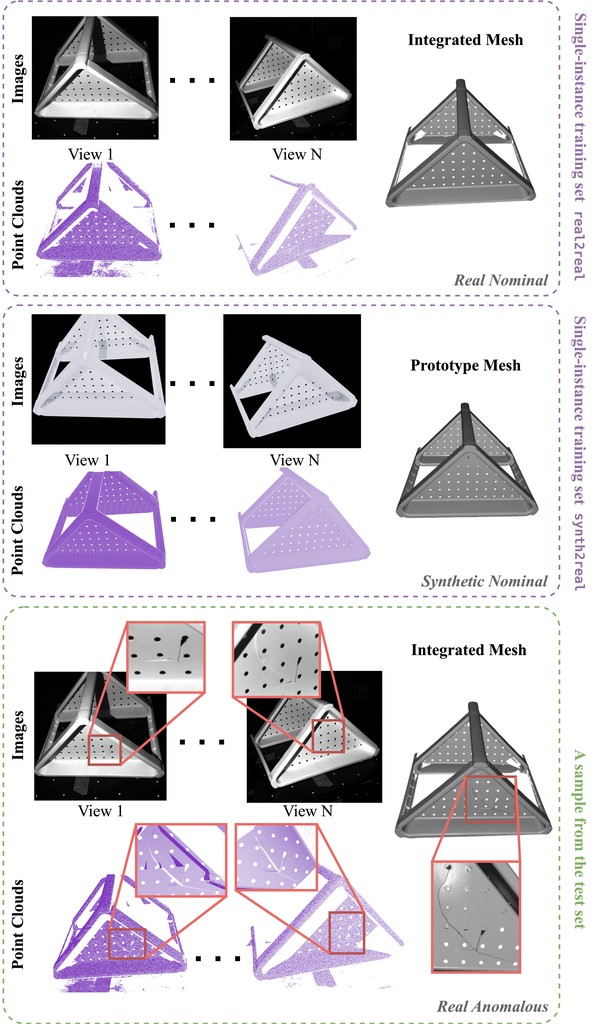}
    \includegraphics[width=0.49\linewidth]{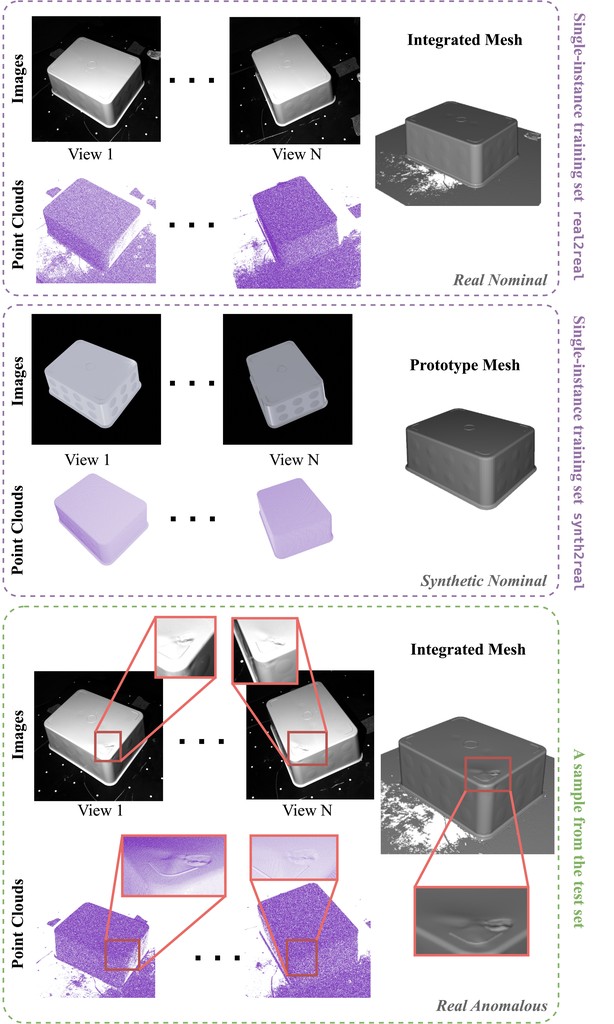}
    \caption{
        \textbf{\dataname{} dataset overview.} 
        From top to bottom: the single-instance real and synthetic training samples for objects \emph{Bathroom Furniture} and \emph{Container}, one of the anomalous samples from the test set.
        }
    \label{fig:overview_b}
\end{figure*}

\begin{figure*}[ht]
    \centering
    \includegraphics[width=0.49\linewidth]{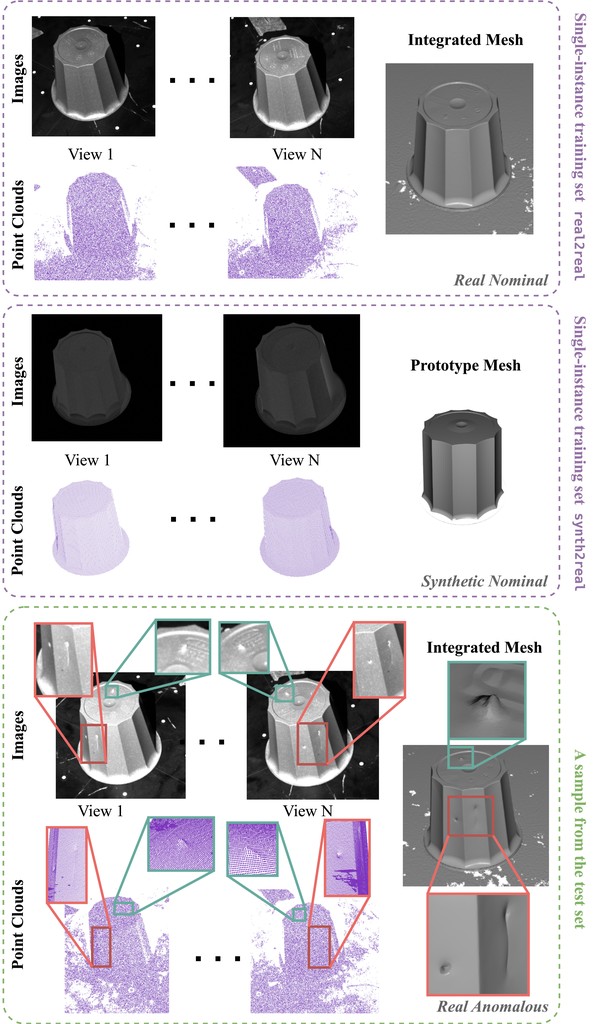}
    \includegraphics[width=0.49\linewidth]{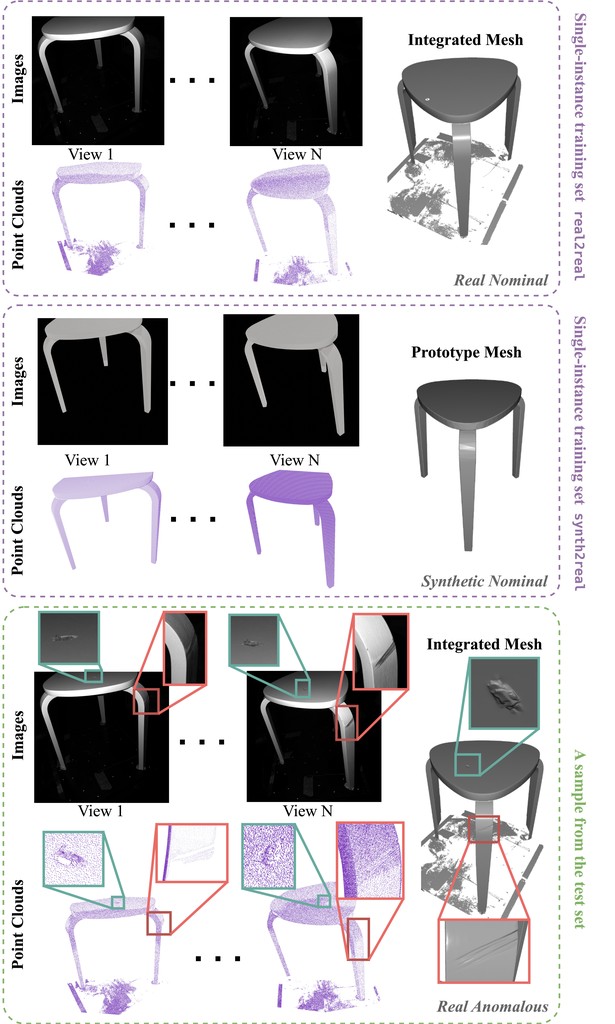}
    \caption{
        \textbf{\dataname{} dataset overview.} 
        From top to bottom: the single-instance real and synthetic training samples for objects \emph{Plastic Vase} and \emph{Wooden Stool}, one of the anomalous samples from the test set.
        }
    \label{fig:overview_c}
\end{figure*}

    \subsection{Ethical statement}
        This research, which was carried out to produce this dataset, adheres to ethical principles and practices in computer vision research. 
        The dataset introduced in this study does not contain any personally identifiable information or sensitive data. 
        It has been collected and processed in a manner that respects individual privacy and avoids potential biases. 
        The dataset and its intended use align with the ethical guidelines outlined by the CVPR community. 
        We have taken care to ensure that the dataset and its potential applications do not pose significant risks to individuals or society.

    \subsection{Additional qualitative results}
        We report in \cref{fig:qualitatives_supp} additional qualitative results concerning the classes of \dataname{} which have not been displayed in in~\cref{fig:qualitatives}.
        We additionally report in \cref{fig:all_multimodal_qual} qualitative results for top-performer multimodal methods reported in \cref{tab:detection_vs_segmentation}.
        
    \subsection{Extended dataset visualizations}
        Akin to \cref{fig:teaser} of the main paper, in \cref{fig:overview_a}, \cref{fig:overview_b}, and \cref{fig:overview_c}, we show training samples, both real and synthetic, as well as defective test samples for other object types present in the \dataname{} dataset. 

    }

\end{document}